\documentclass[conference]{IEEEtran}

\usepackage[dvipsnames, table]{xcolor}
\usepackage[hidelinks, colorlinks=true, citecolor=orange]{hyperref}
\usepackage{graphicx}
\usepackage[font=small]{caption}
\usepackage{array}
\usepackage{multirow}
\usepackage{multicol}
\usepackage{tabularx}
\usepackage{siunitx}
\usepackage{booktabs}
\usepackage{floatrow}
\usepackage{babel,blindtext}
\usepackage{subcaption}
\usepackage{mathtools}
\usepackage{amsmath,amsfonts,amssymb}
\usepackage{algpseudocode}
\usepackage{algorithm}
\usepackage{tcolorbox}
\usepackage{mathrsfs}
\usepackage{listings} 
\usepackage{xspace}
\usepackage[utf8]{inputenc}

\IEEEoverridecommandlockouts                              

\usepackage{amsmath} 

\tcbuselibrary{skins, breakable}

\newtcolorbox[auto counter, number within=section]{promptbox}[2][]{
    breakable,
    colback=gray!10,
    colframe=black,
    fontupper=,
    title={Prompt \thetcbcounter: #2}, 
    label={#1}
}

\newcommand{\du}{\texttt{FACA}\xspace}

\begin{document}

\title{\LARGE \bf FACA: Fair and Agile Multi-Robot Collision Avoidance in Constrained Environments with Dynamic Priorities}

\author{
\IEEEauthorblockN{Jaskirat Singh, Rohan Chandra}
\IEEEauthorblockA{{Dept. of Computer Science, University of Virginia}\\\texttt{\{swd3rm, rohanchandra\}@virginia.edu}}
}

\maketitle


\begin{abstract}
Multi-robot systems are increasingly being used for critical applications such as rescuing injured people, delivering food and medicines, and monitoring key areas. These applications usually involve navigating at high speeds through constrained spaces such as small gaps. Navigating such constrained spaces becomes particularly challenging when the space is crowded with multiple heterogeneous agents all of which have urgent priorities. What makes the problem even harder is that during an active response situation, roles and priorities can quickly change on a dime without informing the other agents. In order to complete missions in such environments, robots must not only be safe, but also agile, able to dodge and change course at a moment's notice. In this paper, we propose \du, a fair and agile collision avoidance approach where robots coordinate their tasks by talking to each other via natural language (just as people do). In \du, robots balance safety with agility via a novel artificial potential field algorithm that creates an automatic ``roundabout'' effect whenever a conflict arises. Our experiments show that \du achieves a improvement in efficiency, completing missions more than \textbf{$3.5\times$} faster than baselines with a time reduction of over \textbf{$70\%$} while maintaining robust safety margins.


\end{abstract}


\section{Introduction}

Multi-agent robots are extensively used in surveillance and security systems \cite{8002674}, logistics \cite{9042827}, delivering products \cite{10113719}, medical emergency and healthcare \cite{lanza2020agents}, motion and traffic analysis, and various other research purposes \cite{alotaibi2016multi, SUDHAKAR20201}. In this work, we focus on the problem of safe, scalable, and efficient heterogeneous multi-robot system navigation in cluttered and constrained environments~\cite{chandra2025multi, chandra2025deadlock}. We consider a distributed (agents are allowed to communicate with each other), but partially observable setting where agents are allowed to communicate with each other. The agents are heterogeneous--they can be UGVs, robots, etc. They have different priorities which can change at any time. The environment consists of various obstacles and small gaps. This problem arises naturally in many crisis situations and presents many issues~\cite{mahadevan2025gamechat, chandra2022gameplan, chandra2022game}. Consider a situation where a surveillance agent, a food delivery agent, and a medical emergency agent are navigating through a post-battle environment for rescue and monitoring. If the environment is congested, we expect the medical emergency agent to be granted the right-of-way whereas the surveillance agent to have the least priority. However, things become complex when these priorities shift without warning~\cite{chandra2023socialmapf}. For example, the surveillance agent might have their priority escalated to assist in an emergency evacuation, making their mission more urgent than food delivery agent.

The key challenge is that in cluttered and constrained environments, many heterogenous multi-robots are unable to manuever smoothly and effectively to accommodate the sudden change in the mission objectives~\cite{zinage2025decentralized, chen2025livepoint, gouru2024livenet}. There are three key ingredients required to facilitate safe, efficient, and scalable navigation in these scenarios: (1) agents must coordinate with one another and keep track of the dynamic priorities in real-time and (2) agents must be agile when maneuvering through small gaps and around obstacles, and (3) motion planning needs to be cheap and fast in order to scale as well as support re-planning during shifting priorities/objectives.

Motion planning research in multi-robot systems has made significant advances in open spaces, but struggle with cluttered environments~\cite{suriyarachchi2022gameopt, suriyarachchi2024gameopt+, chandra2024socialgym}. Conventional local planners often fall short in this regard~\cite{raj2024rethinking, francis2025principles}. For instance, while the standard Artificial Potential Field (APF) method is valued for its simplicity and computational efficiency, but it is notoriously susceptible to the local minima problem, where an agent can become trapped in a concave obstacle configuration having the oscillation, failing to reach its goal.  On the other end of the complexity spectrum, planners like Model Predictive Control (MPC) can generate optimal, collision-free trajectories but impose a significant computational burden that makes them ill-suited for the rapid, on-the-move re-planning required in dynamic, congested environment. This dichotomy highlights a critical gap in the state-of-the-art: existing methods are often either too computationally expensive for rapid re-planning or they struggle to produce trajectories that are simultaneously safe and efficient in dense, constrained scenarios. Furthermore, a related body of work in the trajectory and intent forecasting community~\cite{wu2023intent, chandra2019traphic, chandra2020forecasting, chandra2019robusttp, chandra2020cmetric, chandra2020graphrqi, chandra2022towards, chandra2021using, chandra2020roadtrack, chandra2019densepeds} can take in the past few seconds of history and predict the future trajectories in real-time, however, these methods operate at the trajectory level only. The core research question we ask is: \textit{how does a system of heterogeneous multi-robot systems react quickly to changing priorities and objectives while maneuvering safely through small gaps and around obstacles?}

\noindent\textbf{Main Contributions:} To answer this question, we make two contributions. 
\begin{enumerate}
    \item First, we allow agents to communicate with each other via natural language, much like humans do. We leverage recent advances in large language models (LLMs)~\cite{song2025group, song2025reward, moeini2025survey} to facilitate a dialogue wherein agents update each other about the changing priorities and resolve any conflicts. Our LLM-driven negotiating module has several advantages. First, it can reason about highly complex roles, even ones that are difficult to distinguish. Second, we do not require pre-programming roles apriori or any kind of finetuning or data augmentation as LLMs exhibit good zero-shot reasoning abilities.
    \item Our second contribution consists of a lightweight motion planner using articial potential fields (APFs). APF methods produce agile trajectories quickly, but are known to be susceptible to local minima issues. In this work, we seek to exploit the low computational requirements and agility of APFs while at the same time proposing a solution to the local minima problem. More specifically, we formulate new tangential and radial potential fields that create a ``roundabout'' effect to mitigating local minima.
\end{enumerate}

\section{Related Work}

\subsection{Model-Based Approaches}
Since collision avoidance has been a great topic of interest, several approaches for collision avoidance in mult-robot system have been developed over the years which typically falls under reactive and proactive categories. Most of proactive methods rely on the concept of Velocity Obstacles (VO) introduced by \cite{FS98} which defines the set of all relative velocities that will lead to a collision between a holonomic robot and moving obstacles at some moment in time. VOs offer a tractable alternative to reasoning over inevitable collision states~\cite{ics,ics2} and have been extended to model reciprocal interactions between agents, enabling decentralized multi-agent navigation~\cite{rvo,clearPath}. Subsequent variants handle robots with more complex dynamics~\cite{avo,cco,lqr}, coordinated formations of multiple robots~\cite{coherence,fvo}, and uncertainty in obstacle motion~\cite{pvo,hrvo}. Besides the previously mentioned techniques, that is, Generalized Velocity Obstacles (GVO) was developed by ~\cite{gvo} to define what it means to collide with obstacles in the velocity space and enables a kinematically constrained robot to make a collision-free maneuver by selecting an available control outside of the collision-safe space. Then to deal with the challenges that sampling-based VOs, omnidirectional robot collision avoidance algorithm (ORCA) was recommended ~by \cite{orca}. Since the original work of Van Den Berg et al.~\cite{orca} on holonomic robots, many ORCA-based approaches have been proposed that linearize the VOs or learn such constraints, including approaches for steering differential-drives, car-like robots, and other non-holonomic agents~\cite{rrvo,orcadd,mora1,mora2,jia1}. 


\subsection{Learning-based Approaches}
\label{subsec:reinforcement_learning_based_approahes}

When dynamic environments are taken into consideration the various Deep Reinforcement Learning (DRL) based approaches are widely used for collision avoidance. Building on PPO \cite{schulman2017proximal}, early work learned fully decentralized policies directly from raw 2D lidar \cite{long2018towards,fan2020distributed}, later exploring better lidar state representations \cite{guldenring2020learning}, multi-layout indoor training \cite{perez2020robot}, and late-fusing lidar with camera images \cite{huang2021towards}. To incorporate higher-level interaction cues, researchers feed relative positions/velocities and social signals into the policy \cite{chen2017socially,everett2018motion,everett2021collision}, model human–robot and human–human interactions via attention and relational graphs \cite{chen2019crowd,chen2020relational}, encode crowds with GCNs \cite{chen2020robot}, and use decentralized structural-RNNs \cite{liu2020decentralized}. Yet these agent-level methods often assume open and free spaces and incorpoprtes interaction with the static obstacles; remedies include adding static maps (with dual models for pedestrian presence for safe navigation in dynamic enviornments) \cite{liu2020robot}, late-fusing lidar with pedestrian forecasts \cite{sathyamoorthy2020densecavoid}, and learning lidar latent dynamics unsupervised learning\cite{dugas2020navrep}. Beyond perception, reward shaping has been tackled through knowledge distillation \cite{xu2021human} and square warning regions \cite{pateldwa}, though the former depends on expert data quality and the latter tends to be overly conservative.

\subsection{Multi-Robot Communication through LLMs}

In decentralized, cooperative multi-robot environments, communication methods are needed for coordination ~\cite{chandra2025deadlockfreesafedecentralizedmultirobot}. Researchers are working to integrate LLMs into multi-robot systems (MRS) to address the unique challenges associated with deploying and coordinating MRS \cite{chen2024scalable, mandi2024roco} For example, effective
communication is essential for the MRS to share knowledge, coordinate tasks,and
maintain cohesion in the dynamic environment among individual robots \cite{gielis2022critical}. LLMs
can provide a natural language interface for inter-robot communication, allowing robots to exchange high-level information more intuitively and efficiently instead of predefined communication structures and protocols \cite{mandi2024roco}. The LLMs can understand the mission, divide it into sub-tasks, and assign them to individual robots within
the team based on their capabilities \cite{chen2024emos, liu2025coherent}. The generalization ability across different contexts of LLMs can also allow MRS to adapt to new scenarios without extensive
reprogramming, making them highly flexible during the deployment \cite{wang2024dart,yu2023co}.

\section{Problem Formulation and Background}
\label{problem formulation}

\noindent\textbf{Assumptions:} We assume a distributed system (agents are allowed to communicate) where agents can observe the environment fully, but can only partially observe other agents' states. We further assume agents are self-interested (they optimize their individual objective as opposed to a joint objective), although they are cooperative in the sense that they share information honestly during communication. We also assume agents are point masses (although they can be easily extended to circles).

We begin with a modified Partially Observable Stochastic Game (POSG) \cite{hansen2004dynamic}, defined by the tuple: $\left \langle N, \mathcal{S}, \mathcal{U}, \mathcal{T}\right\rangle$. We assume there are $N$ heterogeneous robots in a confined two-dimensional shared space denoted as $S \in \mathbb{R}^2$ at any given time $t$. A robot's state comprises of its position, $s_i(t) = \bigl(x_i(t), y_i(t)\bigl)$ and turning angle $\theta_i(t) \in \mathcal{U}$, $0 \leq \theta_i \leq 2\pi$. A robot's action at time $t$ is expressed as velocity $v_i(t)\in\mathbb{R}^2$ and $0 \leq v_i(t) \leq \bar{v_i}$, where $\bar{v_i}$ is the maximum velocity. Each robot has a goal $g_i \in \mathcal{S}$. The discrete transition function is given as $    s_{i+1}(t) = s_i(t) + v_i(t)\cos(\theta_i(t))$.
Initially each robot is assigned a priority $\rho_i(t) \in \mathbb{R}$ that may change based on various situations, resulting in the assignment of a new priority at $t+1$. The objective for each robot is to reach its goal as fast as possible. Thus, each robot $i$ solves the following:
\begin{subequations}
\begin{align}
\arg\min_{\mathcal{U}} \quad & \text{Time to Goal}_i \label{eq:obj} \\
\text{s.t.} \quad 
& \left \lVert s_i(t) - s_j(t)\right \rVert &\geq \text{safe distance} \label{eq:constraint1} \\
& \left \lVert s_i(t) - \text{obstacle}\right \rVert &\geq \text{safe distance} \label{eq:constraint2} \\
& \text{TTG}_i < \text{TTG}_j \ &\text{if} \ \rho_i > \rho_j \label{eq:constraint3}
\end{align}
\label{eq: opt}
\end{subequations}

\noindent Constraint~\eqref{eq:constraint1} operates over all other robots $j$. Constraint~\eqref{eq:constraint2} operates over all obstacles. All constraints hold for some fixed horizon in the future. Constraint~\eqref{eq:constraint3} operates over all pairs of agents $i,j$ and measures fairness where TTG$_i$ represents the time to goal for robot $i$. A fair outcome is one where, in cases of conflict, robots with higher priority are expected to be given the right of way, therefore reaching the goal first.

\begin{figure*}
    \centering
    \includegraphics[width=\textwidth]{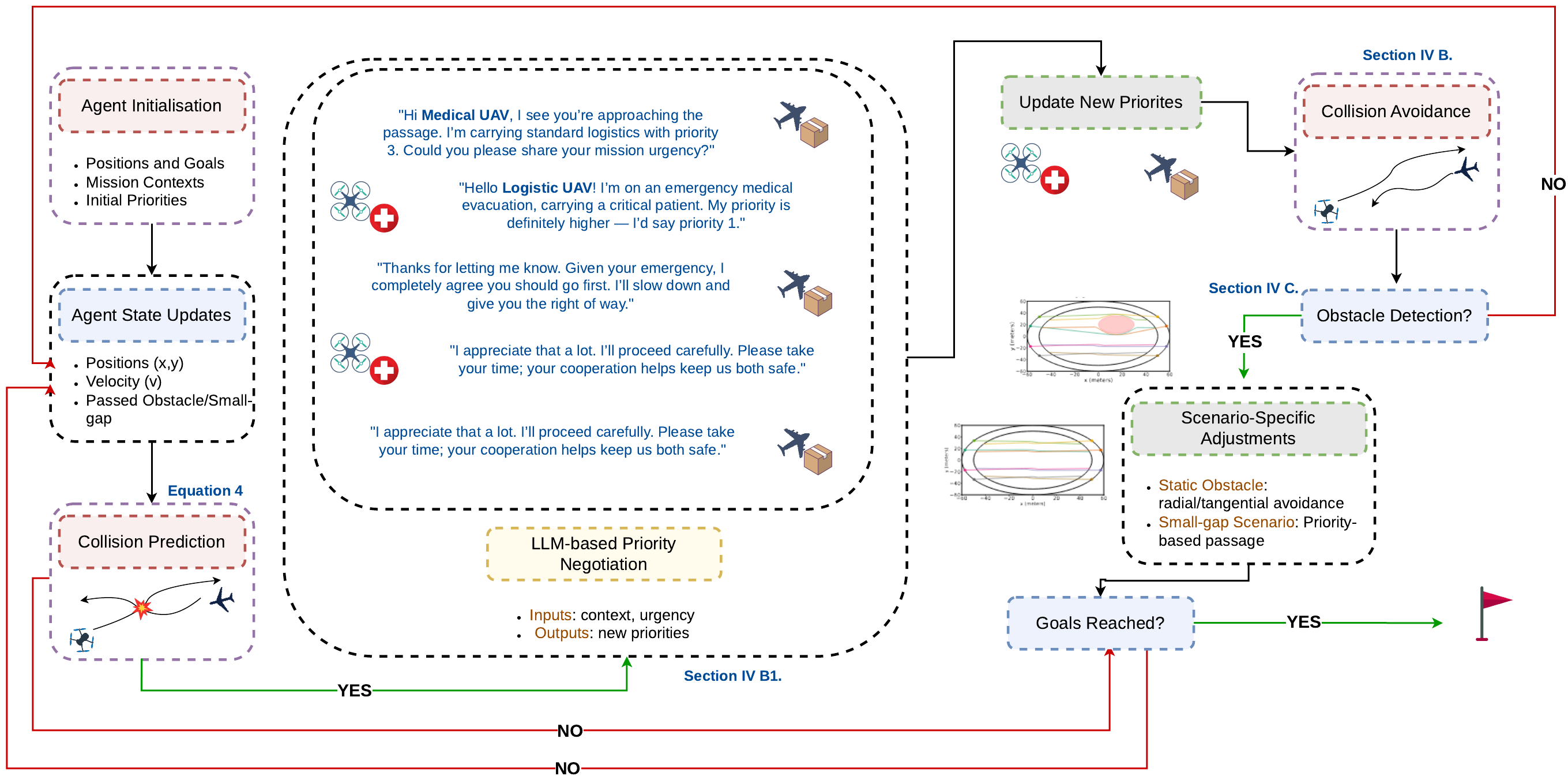}
    \caption{An overview of the proposed framework for autonomous robot collision avoidance using Large Language Model (LLM) based negotiation in heterogenous robots. The system operates in a loop, beginning with agent initialization. Upon predicting a potential collision, an LLM is used to dynamically negotiate priorities based on mission context. As illustrated in the \textit{LLM-based Priority Negotiation} box where the      ``Medical robot'' is granted higher priority over a ``Logistic robot'' with inputs as mission, and urgency to output new, dynamically-assigned priorities. The framework then applies scenario-specific adjustments for challenges like obstacle or small-gap passages before executing the final collision avoidance maneuvers.}
            \label{fig:overview}
            \vspace{-10pt}
\end{figure*}

\subsection*{Artificial Potential Fields}
We build our algorithm using artificial potential fields \cite{khatib} which define an attractive potential $(\mathcal{U}_i^A)$ towards the goal and a repulsive potential $(\mathcal{U}_i^R)$ from the obstacles for collision avoidance. These are then combined to the obtain the net potential $(\mathcal{U}_i^N) = \mathcal{U}_i^A + \mathcal{U}_i^R$.
The corresponding forces $\mathcal{F}_i^N$, $\mathcal{F}_i^A$, and $\mathcal{F}_i^R$ are then derived by taking the negative gradient of the respective potential functions.
The standard artificial potential field implementation uses the following repulsive and attraction potential functions:
        {\begin{eqnarray}
            \mathcal{U}_i^A &=& \frac{1}{2}\eta^A\left \lVert g_i - s_i\right \rVert^2,\label{eq:attraction}\\
            \mathcal{U}_i^R(j) &=& \frac{1}{2}\eta^R\left(\frac{1}{\left \lVert s_i - s_j\right \rVert} - \frac{1}{\sigma}\right )^2, \label{eq:repulsion}
        \end{eqnarray}}
where $\mathcal{U}_i^R(j)$ is the potential experienced by $i$ due to $j$ and $||\cdot ||$ is the euclidean distance between robots $i$ and $j$ or between the robot and the goal, $\sigma$ is the safety distance, and $\eta^A, \eta^R>0$, represent the attractive and repulsive gain respectively.  

A key limitation of the APF method is that the classical APF method when directly applied to different settings often gets stuck in local minima~\cite{khatib}. Therefore, it needs to be adapted and re-designed for the particular application. 
Therefore, we introduce a new APF collision avoidance algorithm that mitigates the local minima problem in multi-robot navigation via a smooth roundabout effect.


\section{\du Framework}\label{decision-making framework}
Here, we describe $\du$ in detail. First, we review the technical setup of our environment and provide an overview of our approach. Next, we present our implementation of LLM‑based priority determination and the agent‑to‑agent dialogue negotiation mechanism for task prioritization. Finally, we explain our collision‑avoidance strategies for robots, including navigation around obstacle and through confined spaces such as small gaps.

\subsection{Technical Approach}

The overview of our proposed technical approach is shown in Figure \ref{fig:overview}. At a high level, at each discrete time step $t$, each robot at $s_i(t)$ updates its position to $s_i(t+1)$ by computing a new velocity $v_i(t+1)$ while solving the optimization problem in~\eqref{eq: opt}. The optimization proceeds as follows: every robot observe the environment and checks for potential collisions via Equation~\eqref{eq: collision_check}. If a collision is detected, then two things happen. First, the robots engage in an LLM-based dialogue (Section~\ref{LLMs for Priority Determination}) to exchange mission urgency and context information, dynamically adjusting their priorities. Simultaneously, the robots execute the newly proposed ``Roundabout Effect'' collision avoidance algorithm (Section~\ref{eq:attraction} and~\ref{eq:repulsion}), which in turn uses the latest priorities, guiding robots to yield or proceed based on negotiated right-of-way. The collision avoidance routine determines a new velocity to execute via Equation~\eqref{eq: velocity}.




\subsection{Collision Avoidance Around Dynamic Agents: The ``Roundabout Effect'' (Solving Constraint~\ref{eq:constraint1})}
\label{subsec: roundabout}

We divided our approach into two parts to avoid collision with other robots. First, we predict the collision by considering a 2D engagement scenario with two robots, $i$, and $j$. Let the initial position of $i$ be $s_{i}(t) = (x_{i}, y_{i})$, and that of $j$ be $s_{j}(t) = (x_{j}, y_{j})$. The robots have velocities $v_{i}(t)$, and $v_{j}(t)$, respectively. We calculate the distance of closest approach, $\Sigma(t)$ as:
{\begin{equation}
        \begin{split}
            \Sigma(t)^2 = \bigl(s_{i}(t) - s_{j}(t)\bigl)\cdot\bigl(s_{i}(t) - s_{j}(t)\bigl)\\ + 2\bigl(s_{i}(t) - s_{j}(t)\bigl)\cdot\bigl(v_{i}(t)-v_{j}(t)\bigl) t_{col}\\+ \bigl(v_{i}(t)-v_{j}(t)\bigl)\cdot\bigl(v_{i}(t)-v_{j}(t)\bigl)t_{col}^2
        \end{split}
        \label{eq: collision_check}
        \end{equation}}
        

Next,
If $\Sigma(t)$ is less than a safe distance threshold, that means it will (1) trigger the LLM-based dialogue communication for priority negotiation as explained in \ref{LLMs for Priority Determination} and (2) perform collision avoidance maneuver as described in Sections~\ref{eq:attraction} and~\ref{subsubsec: repulsion}: 
\subsubsection{LLMs for Priority Determination and Negotiation (Solving Constraint~\ref{eq:constraint3})}
\label{LLMs for Priority Determination}

The main goal of LLM-based dialogue is to satisfy Constraint~\eqref{eq:constraint3} by ensuring that higher priority robots are given the right of way in case of a potential collision. Each agent has access to an LLM (gpt-4o-mini~\cite{openai2024gpt4ocard})
instance through the OpenAI API to perform context-aware, adaptive priority negotiation in real-time, and has an initial prompt which we provide in supplementary material.
When the agents observe each other, they take turns sending messages; they receive the last message from the other agent, add it to the dialogue history, and query the
LLM for a reply to send back. The robots keep moving around the environment while communicating and will stop conversing if they reach a consensus on which agent's task has the higher priority. If the consensus leads to a change in the priorities, the new priorities are updated.
Additionally, we considered complex scenarios where negotiating robots perform critical operations with similar priorities. As shown in Figure~\ref{lst:dialogue_example_1}, a representative case involves \textit{robot i} transporting a patient to a city hospital while \textit{robot j} transports a ventilator to the same hospital. With both missions assigned the highest priority, the robots negotiate to reach a consensus. 

\begin{promptbox}[lst:dialogue_example_1]{Example LLM-Mediated Dialogue}

\textbf{robot $i$:} robot $j$, conflict warning. I am transporting a critical patient to City General Hospital, my mission is top priority. I must maintain my current heading.

\par\medskip 

\textbf{robot $j$:} Acknowledged, robot $i$. I am also on a critical mission, transporting time-sensitive medical equipment for emergency surgery. My priority is also maximal.

\par\medskip

\textbf{robot $i$:} Understood. My patient is unstable, and my distance to the hospital is 3.1km. What is your distance to goal?

\par\medskip

\textbf{robot $j$:} My distance to City General is 5.8km. Your patient's immediate proximity is more critical. I will yield. I am altering my course and reducing speed. Proceed safely.

\par\medskip

\textbf{Agreement (LLM Output):} \texttt{"\{$i$: high priority, $j$: low priority\}"}

\end{promptbox}


\subsubsection{{Non-linear Exponential Attractive Force}}
\label{subsubsec: attraction}
One of the issues with the attractive field given in Equation \eqref{eq:attraction} is that the attraction force strength reduces the closer a robot gets to its goal. If there is an obstacle near the goal, the repulsive field from the obstacle overpowers the attractive field and it results in a local minima in the potential that is away from the goal. Therefore, we proposed the following attractive force. We define the potential function $\mathcal{U}_A$ as:

{\begin{equation}
    \mathcal{U}_A(d) = \kappa^A d - \kappa^A \frac{\sqrt{\pi}}{2\sqrt{-\varphi^A}} \text{erf}(\sqrt{-\varphi^A}d)
\end{equation}}

\noindent where $d$ is the distance between the goal and current position of the robot and $\varphi^A < 0$ represents the attractive field spread. Now, $ \mathcal{F}^A = -\nabla \mathcal{U}_A(d) = -\left(\frac{\partial\mathcal{U}_A}{\partial d}\right) \widehat d$ where $\widehat d  = \frac{s_i -  g_i}{\left \lVert s_i - g_i\right \rVert}$. Using the standard derivative $\frac{\partial}{\partial z}\text{erf}(z) = \frac{2}{\sqrt{\pi}}e^{-z^2}$:




{\begin{equation*}
\begin{aligned}
    \frac{\partial \mathcal{U}_A}{\partial d} &= \frac{\partial}{\partial d}\left[ \kappa^A d - \kappa^A \frac{\sqrt{\pi}}{2\sqrt{-\varphi^A}} \text{erf}(\sqrt{-\varphi^A}d) \right] \\
    &= \kappa^A - \kappa^A \frac{\sqrt{\pi}}{2\sqrt{-\varphi^A}} \cdot \left(\frac{2}{\sqrt{\pi}}e^{-(\sqrt{-\varphi^A}d)^2}\right) \cdot \sqrt{-\varphi^A} \\
    &= \kappa^A - \kappa^A e^{-(-\varphi^A d^2)} \\
    &= \kappa^A \left(1 - e^{\varphi^A d^2}\right)\\
    &= \kappa^A \Bigl(1-\exp\bigl(\varphi^A\left \lVert g_i - s_i\right \rVert^2\bigl)\Bigl)
\end{aligned}
\end{equation*}}
This term represents the {magnitude} of the attractive force. Then,


\begin{equation}
\begin{aligned}
\mathcal{F}^A &= -\left(\frac{\partial\mathcal{U}_A}{\partial d}\right)\widehat d \\[6pt]
&= -\kappa^A 
\Bigl(1-\exp\bigl(\varphi^A\lVert s_i 
- g_i\rVert^2\bigr)\Bigr) \\
&\quad \times \biggl[\frac{ s_i - g_i}{\lVert  s_i - g_i \rVert}\biggr]
\end{aligned}
\end{equation}



where $\kappa^A > 0$ represents the attractive gain, which is used to adjust the intensity of the attractive field. 

\subsubsection{{Tangential Exponential Repulsive Force}}
\label{subsubsec: repulsion}

Let $d_{ij} = \lVert s_i - s_j \rVert$ denote the distance between robot $i$ and robot $j$, and let $\widehat d_{ij} = \dfrac{s_i - s_j}{d_{ij}}$ denote the corresponding unit vector. We define the repulsive potential as

\begin{equation}
    \mathcal{U}_R(d_{ij})
    = -\,\kappa^R \frac{\sqrt{\pi}}{2\sqrt{\varphi^R}}
    \,\operatorname{erf}\!\bigl(\sqrt{\varphi^R}\,d_{ij}\bigr),
\end{equation}

\noindent where $\kappa^R > 0$ represents the repulsive gain, which is used to adjust the intensity of the repulsive field, and $\varphi^R > 0$ represents the repulsive field spread. Now, $\mathcal{F}_{i}^{R} = -\nabla \mathcal{U}_R(d_{ij}) = -\Bigl(\frac{\partial \mathcal{U}_R}{\partial d_{ij}}\Bigr)\widehat d_{ij}$. Using the standard derivative $\frac{\partial}{\partial z}\text{erf}(z) = \frac{2}{\sqrt{\pi}}e^{-z^2}$:

\begin{equation*}
\begin{aligned}
    \frac{\partial \mathcal{U}_R}{\partial d_{ij}}
    &= \frac{\partial}{\partial d_{ij}}
    \left[-\,\kappa^R \frac{\sqrt{\pi}}{2\sqrt{\varphi^R}}
    \,\operatorname{erf}\!\bigl(\sqrt{\varphi^R}\,d_{ij}\bigr)\right] \\[6pt]
    &= -\,\kappa^R \frac{\sqrt{\pi}}{2\sqrt{\varphi^R}}
    \cdot \left(\frac{2}{\sqrt{\pi}} e^{-(\sqrt{\varphi^R}d_{ij})^{2}}\right) \cdot \sqrt{\varphi^R} \\[6pt]
    &= -\,\kappa^R e^{-\varphi^R d_{ij}^{2}}.
\end{aligned}
\end{equation*}

This term represents the slope of the repulsive potential. Then,

\begin{equation}
\begin{aligned}
    \mathcal{F}_{i}^{R}
    &= -\left(\frac{\partial \mathcal{U}_R}{\partial d_{ij}}\right)\widehat d_{ij} \\[6pt]
    &= -\left(-\,\kappa^R e^{-\varphi^R d_{ij}^{2}}\right)\widehat d_{ij} \\[6pt]
    &= \kappa^R e^{-\varphi^R \lVert s_i - s_j\rVert^{2}}
    \left[\frac{s_i - s_j}{\lVert s_i - s_j\rVert}\right].
\end{aligned}
\end{equation}

To prevent head-on standstills and induce a roundabout effect, the repulsive force is rotated by ninety degrees using the rotation operator ${\small \xi=\begin{bmatrix}0 & 1\\-1 &0\end{bmatrix}}$. The resulting tangential exponential repulsive force is

\begin{equation}
\begin{aligned}
    \mathcal{F}_{i}^{R}
    &= \xi\,\mathcal{F}_{i}^{R,\text{rad}} \\[6pt]
    &= \kappa^R
    \exp\bigl(-\varphi^R \lVert s_i - s_j\rVert^{2}\bigr)\,
    \xi \left[\frac{s_i - s_j}{\lVert s_i - s_j\rVert}\right].
\end{aligned}
\end{equation}


To incorporate priority obtained from the LLM-based dialog discussed in Section~\ref{LLMs for Priority Determination} into the repulsive force, a scaling factor $\frac{\rho_i}{\rho_i}$ is used. When two robots $i$ and $j$ possess equal priorities, their interaction resembles vehicles navigating a traffic circle, continuously maneuvering around each other to prevent collisions. However, if $i$ is assigned a higher priority mission relative to $j$ (i.e., $\rho_i > \rho_j$), the repulsive force exerted on $j$ by $i$ will be greater compared to the force experienced by $i$ from $j$. Consequently, $j$ will undergo a greater deviation from its path than $i$, allowing $i$ to maintain a more direct trajectory towards its objective. The resulting repulsion force is the modified as:

{\begin{equation}
    \mathcal{F}_i^R = \frac{\rho_{i}}{\rho_{j}}\kappa^R \exp\bigl(-\varphi^R\left \lVert s_i - s_j\right \rVert^2\bigl))\,
    \xi \left[\frac{s_i - s_j}{\lVert s_i - s_j\rVert}\right]
\end{equation}}


The rotation leads to the creation of the new heading angle $\theta_i(t+1)$ which is simply the angle of this vector as described below:
{\begin{equation}
    \theta_i(t+1) = \tan^{-1}\left(\mathcal{F}_i^A + \mathcal{F}_i^R\right)
\end{equation}}


\subsection{Collision Avoidance Around Static Obstacles (Solving Constraint~\ref{eq:constraint2})}
\label{subsec: obstacle}

Our approach to collision avoidance for static obstacles mirrors that of Section~\ref{subsec: roundabout} with a key difference. Namely, assuming a circular obstacle with radius  $r$ and center $\mu$, the repulsive force $\mathcal{F}^R_i$ is no longer perpendicular to the direction of the robot towards the obstacle. Instead, $\mathcal{F}^R_i$, is less conservative, and is along the tangent direction to the boundary of the obstacle. Formally, let $\phi$ be defined as the rotation matrix required to turn $s_i - \mu$ so that it becomes the tangent to the obstacle via simple geometry. Then,

\begin{equation}
    \mathcal{F}^R_o = \kappa^R \Bigl(\exp\bigl(-\varphi^R\left \lVert s_i - \mu - r\right \rVert^2\bigl)\Bigl)\phi
\end{equation}

\noindent The term $\exp\bigl(-\varphi^R\left \lVert s_i - \mu - r\right \rVert^2\bigl)$ denotes the strength of the force depending on how far the robot is from the nearest point on the boundary of the obstacle. Then, the turning angle can be computed as,

{\begin{equation}
    \theta_i(t+1) = \tan^{-1}\left(\mathcal{F}_i^A + \mathcal{F}_o^R\right)
\end{equation}}




\subsection{Final Velocity Computation using $\theta_i$}
Once the turning angle is determined, it rotates the current velocity vector towards the selected avoidance direction by angle $\theta$ to obtain a new velocity $v_{i}(t+1)$ as shown below.
\begin{equation}
v_i(t+1) = 
\begin{bmatrix}
\cos\theta_i & -\sin\theta_i \\
\sin\theta_i & \cos\theta_i
\end{bmatrix}
v_i(t)
\label{eq: velocity}
\end{equation}

To ensure smoothness, we blend the robot's current velocity and the avoidance direction, which is modified by adjusting by calculated turning angle. This blending is done using a factor that increases as the robot gets closer to the obstacle, to smoothly transition from goal-directed movement to avoidance behavior. In order to perform a progressive change in behavior as the robot gets closer to the obstacle, the blending factor $(\beta)$ is determined which is dependent on the robot's distance from the obstacle boundary. The equation representing this mechanism has been described below: 

{\begin{equation}
    v_i(t+1) \gets \beta v(t) + (1-\beta)v_{i}(t+1)
\end{equation}}


\noindent\textit{Remark:} To maneuver through small gaps, we use the concept of subgoals. When a robot $i$ starts initially from their respective positions, their motion is attracted toward the center of the gap which acts as subgoal which we denote as $\omega$. The attractive acceleration towards this subgoal is ${\Phi}_i^{\text{attr}} = \kappa^A \left(1 - e^{-c \left \lVert{\omega} - {s}_i\right \rVert^2}\right) \frac{ {s}_i - {\omega}}{\left \lVert {s}_i - {\omega}\right \rVert}$ where \( c \) controls the rate of exponential decay. Once $i$, crosses the wall, it changes its goal from the small gap to its final goal \(  g_i \). The repulsion forces are the same as used in Section~\ref{subsubsec: repulsion}.



\section{Experiments and Results} 
\label{simulation and results}

We address the following questions:
\begin{itemize}
    \item \textit{Q1. What are the benefits of the LLM when negotiating consensus between multiple heterogeneous robots?}
    \item \textit{Q2. How does the proposed collision avoidance framework compare to traditional APF and MPC-based baselines?}
    \item \textit{Q3. How challenging are constraints? Why should we care about constraints?}
\end{itemize}


\subsection{Simulation Environment and Scenario Setup}

The simulation is setup where $n=4$ and $n=8$ robots are initialized with starting and goal positions arranged in a symmetrical pattern on a $\frac{\pi}{16}$ radian arc on a circle with $50$ meters radius. Each robot must travel to its goal placed antipodal with respect to its starting position on the arena boundary. The priority for each robot was drawn from a normal distribution with $\mu = 3$ and $\sigma = 1$. Initially assigned priorities gets dynamically changed to another set of priorities in the middle of a simulation. We generate and add two different sources of complexity. The first spatial complexity consists of a circular shaped obstacle with a fixed radius placed in the arena for creating a more dense and complex environment. The second complexity consists of a small gap between two vertical walls placed at $x=0$. Simulations were performed in Python in 8-core AMD Ryzen™ 9 5900HS CPU and GeForce RTX™ 3060 GPU. Parameter values used in simulation are provided in the supplementary material. Each scenario is averaged across 100 simulations with a different random seeds.

\subsection{Evaluation Criteria and Baselines} 
\label{Evaluation Criteria}

First, we measure the Time to Goal (TTG) to evaluate the overall time taken by the robots to reach their goal. Next, we would like to measure how much distance each robot maintains form other robots, the obstacle, and the walls. That is, the robots should not aim to shoot towards their destination at the risk of colliding or cutting closer to other objects. So we also measure the Mean Minimum Distance (MMD) between all the agents. Next, we measure the Flow Rate (FR) which measures the flow through a gap where multiple agents are trying to get through. It is defined as: $FR = \frac{N}{zT}$ where $N$ is the number of agents, $T$ is the makespan (in seconds), while $z$ is the gap width (in unit meters). Flow rate gives a metric quantifying the overall ``volume of agents'' successfully navigating a constrained space.

Our baselines consists of the classical APF method utilizing the attraction and repulsion forces given by Equations~\eqref{eq:attraction} and~\eqref{eq:repulsion}, respectively, and an MPC baseline, to demonstrate the collision avoidance and navigation capabilities of our method. We also conduct an ablation study where we probe each component of our approach in isolation. Finally, we also compare our approach with an existing state-of-the-art multi-robot method, RIPNA~\cite{George2009ARI}, which implements multi-robot navigation but without priority handling.
        


\begin{figure*}[h!] 
    \centering 
    
    \begin{minipage}{0.32\textwidth}
        \centering
        \includegraphics[width=\linewidth]{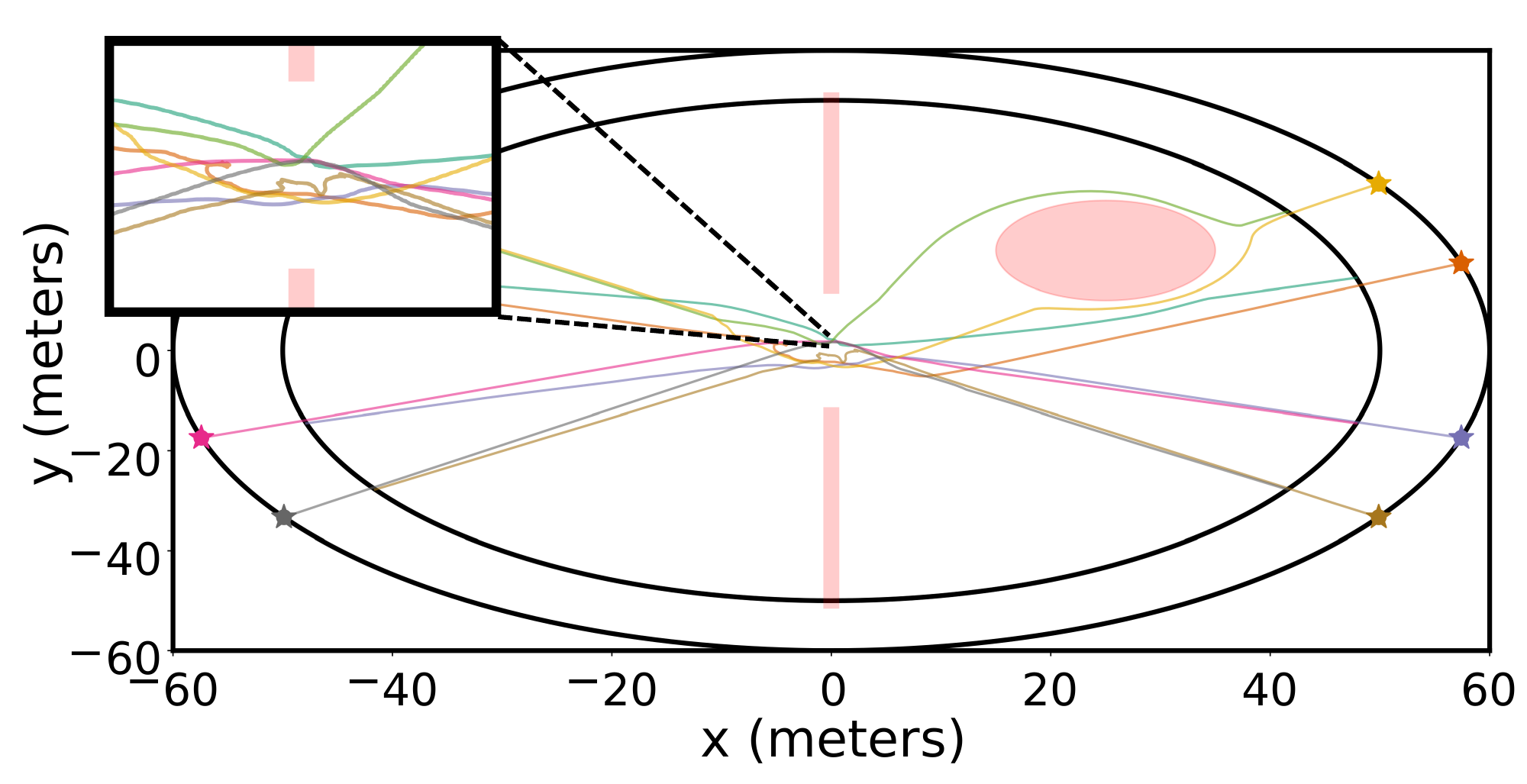}
        \subcaption{\du}
        \label{fig: du}
    \end{minipage}\hfill 
    \begin{minipage}{0.32\textwidth}
        \centering
        \includegraphics[width=\linewidth]{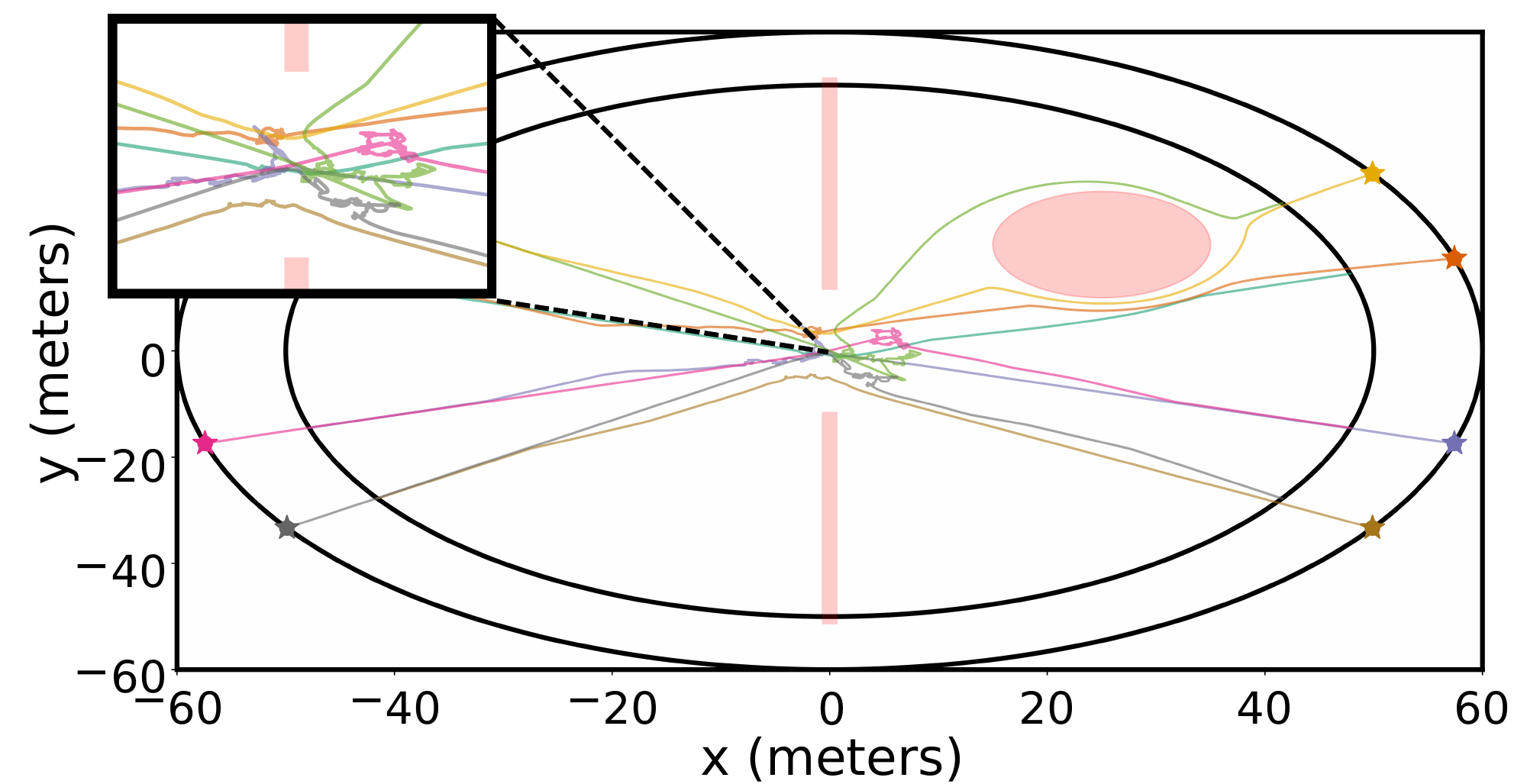}
        \subcaption{Classical APF}
        \label{fig: apf}
    \end{minipage}\hfill
    \begin{minipage}{0.32\textwidth}
        \centering
        \includegraphics[width=\linewidth]{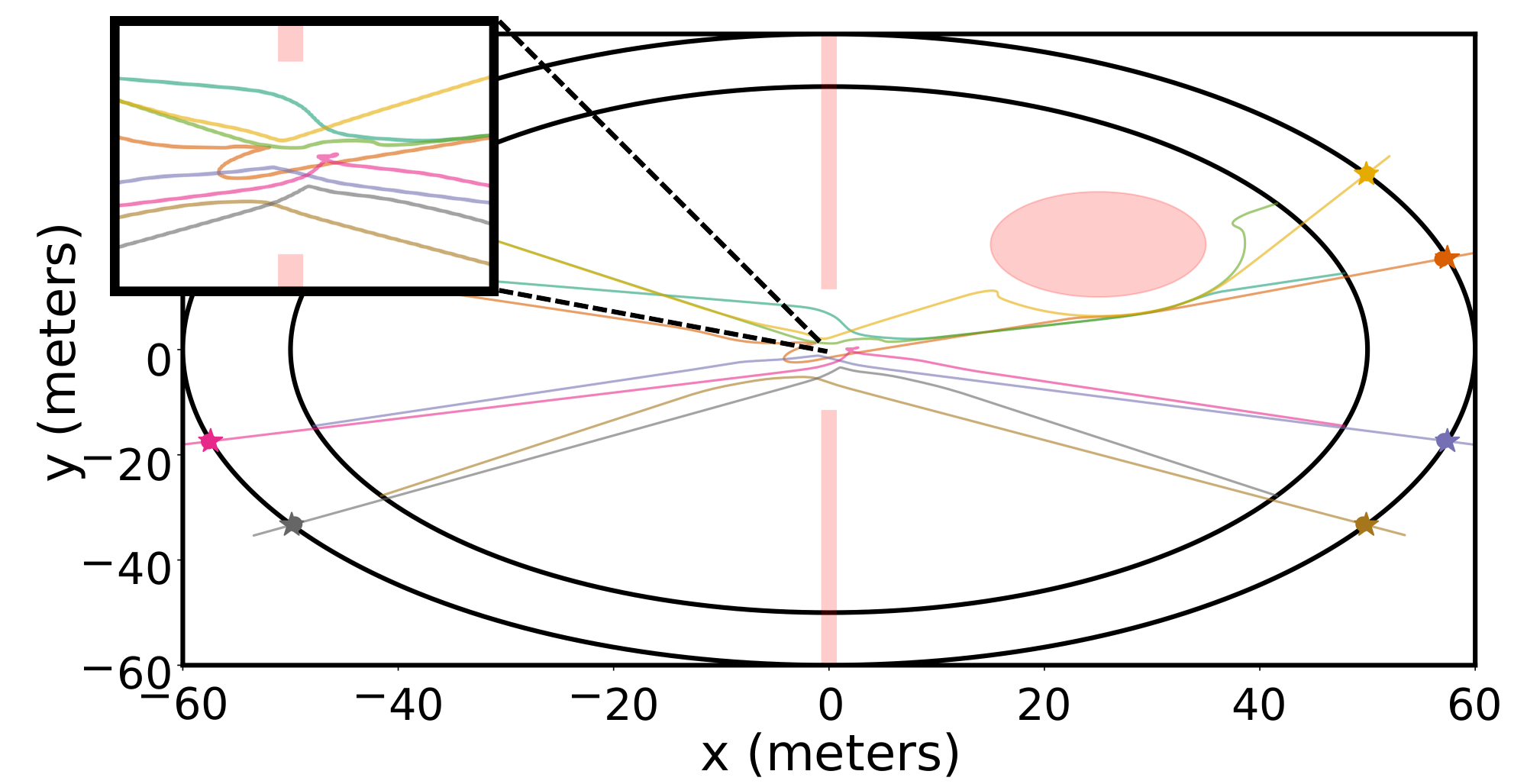} 
        \subcaption{MPC}
        \label{fig: mpc}
    \end{minipage}
    
    \caption{Comparing \du with baseline collision avoidance methods.}
    \label{fig:three_images}
\end{figure*}


\begin{table}[t]
\centering
\resizebox{\columnwidth}{!}{\begin{tabular}{@{}lcccccc@{}} 
\toprule
Scenarios & \multicolumn{2}{c}{TTG (s)} & \multicolumn{2}{c}{MMD (m)} & \multicolumn{2}{c}{FR} \\
\cmidrule(lr){2-3} \cmidrule(lr){4-5} \cmidrule(lr){6-7}
& $n=4$ & $n=8$ & $n=4$ & $n=8$ & $n=4$ & $n=8$ \\
\midrule
No LLM, No Constraints      & 8.79 & 10.02 & 1.98 & 1.49 & - & -\\
With LLM, No Constraints    & {2.51} & {2.86} & 1.95 & 1.47 & - & - \\
With LLM, obstacle only          & {8.24} & {9.63} & 1.70 & 1.33 & - & -\\
With LLM, Small Gap only    & {2.96} & {3.00} & 0.17 & 0.21 & 2.70 & 5.00\\
\bottomrule
\end{tabular}%
}
\caption{Evaluating \du in cases with varying spatial constraints with 4 and 8 agents. Performance is evaluated based on the Time-to-Goal (TTG) in seconds and the Mean Minimum Distance (MMD) in meters. FR is measured at small gaps, hence only shown for \textit{row 4}.} 
\label{tab:experiment_iesults_equal}
\end{table}

\begin{figure*}
  \begin{subfigure}{0.32\textwidth}
    \includegraphics[width=\linewidth]{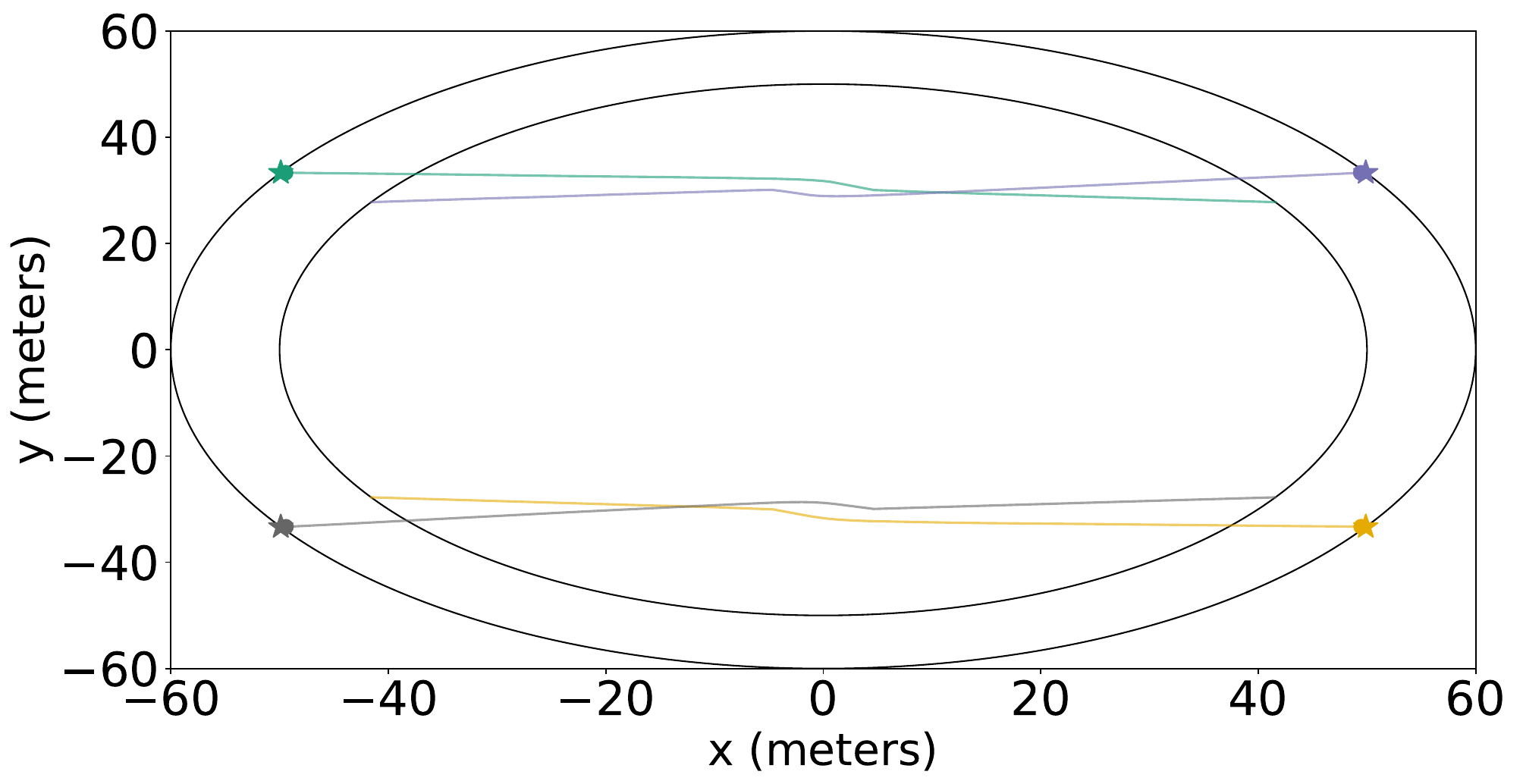}
    \caption{Free space with $n=4$.}
    \label{fig:figure1}
  \end{subfigure}%
  \hfill
  \begin{subfigure}{0.32\textwidth}
    \includegraphics[width=\linewidth]{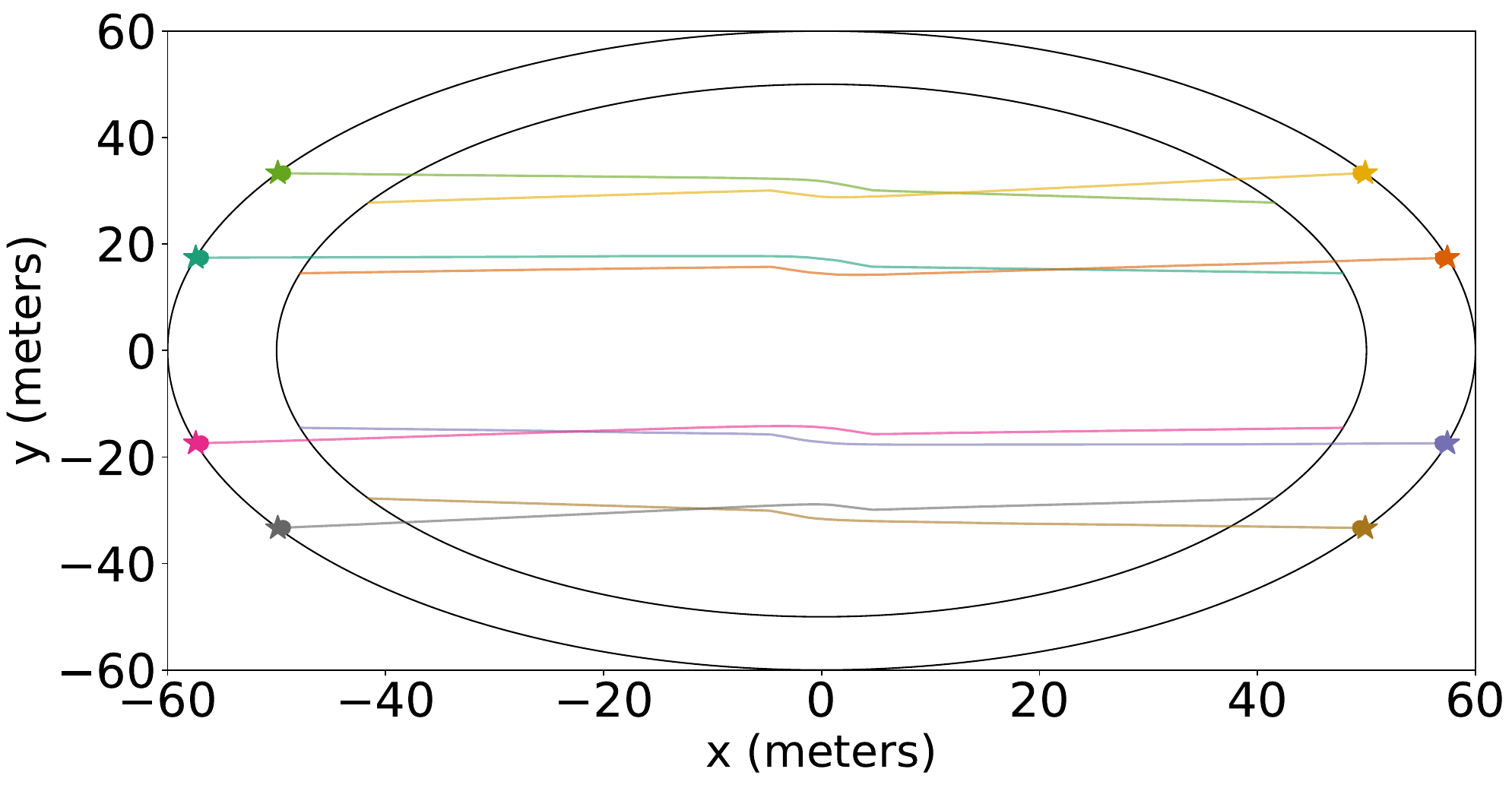}
    \caption{Free space with $n=8$.}
    \label{fig:figure2}
  \end{subfigure}
  \hfill
  \begin{subfigure}{0.32\textwidth}
    \includegraphics[width=\linewidth]{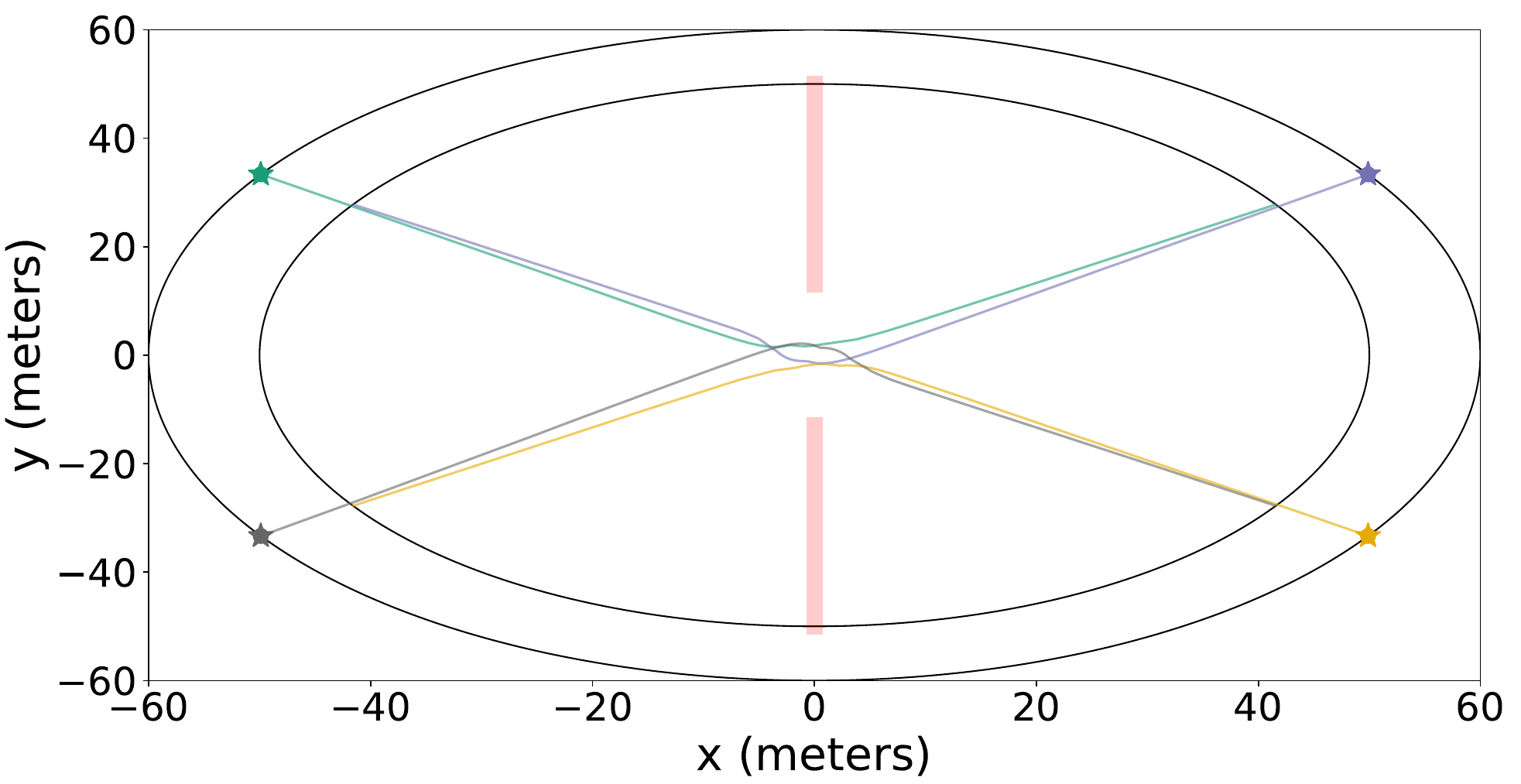}
    \caption{Small gap with $n=4$.}
    \label{fig:figure3}
  \end{subfigure}
  \hfill
  \begin{subfigure}{0.3\textwidth}
    \includegraphics[width=\linewidth]{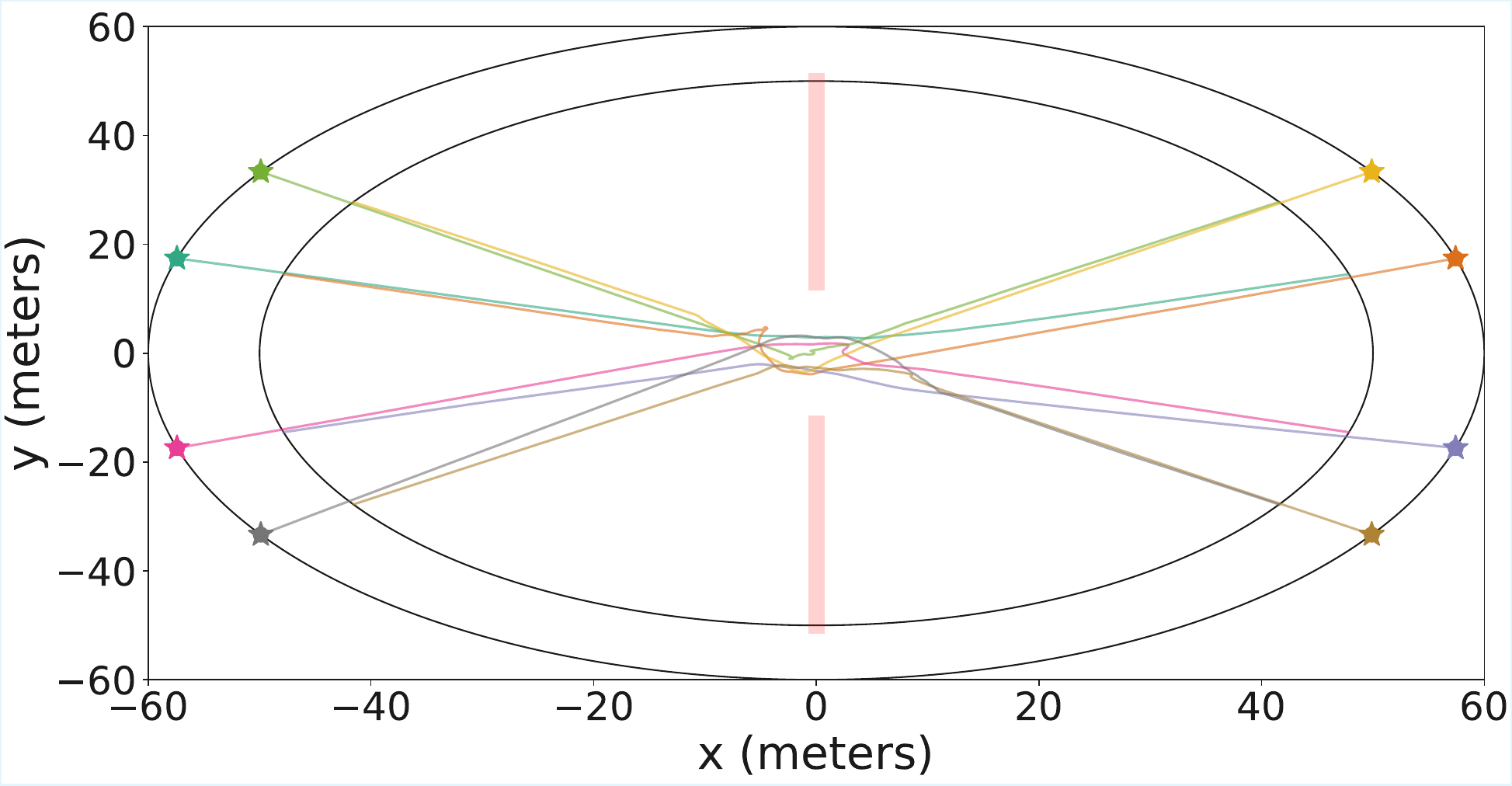}
    \caption{Small gap with $N=8$.}
    \label{fig:figure4}
  \end{subfigure}
  \hfill
  \begin{subfigure}{0.3\textwidth}
    \includegraphics[width=\linewidth]{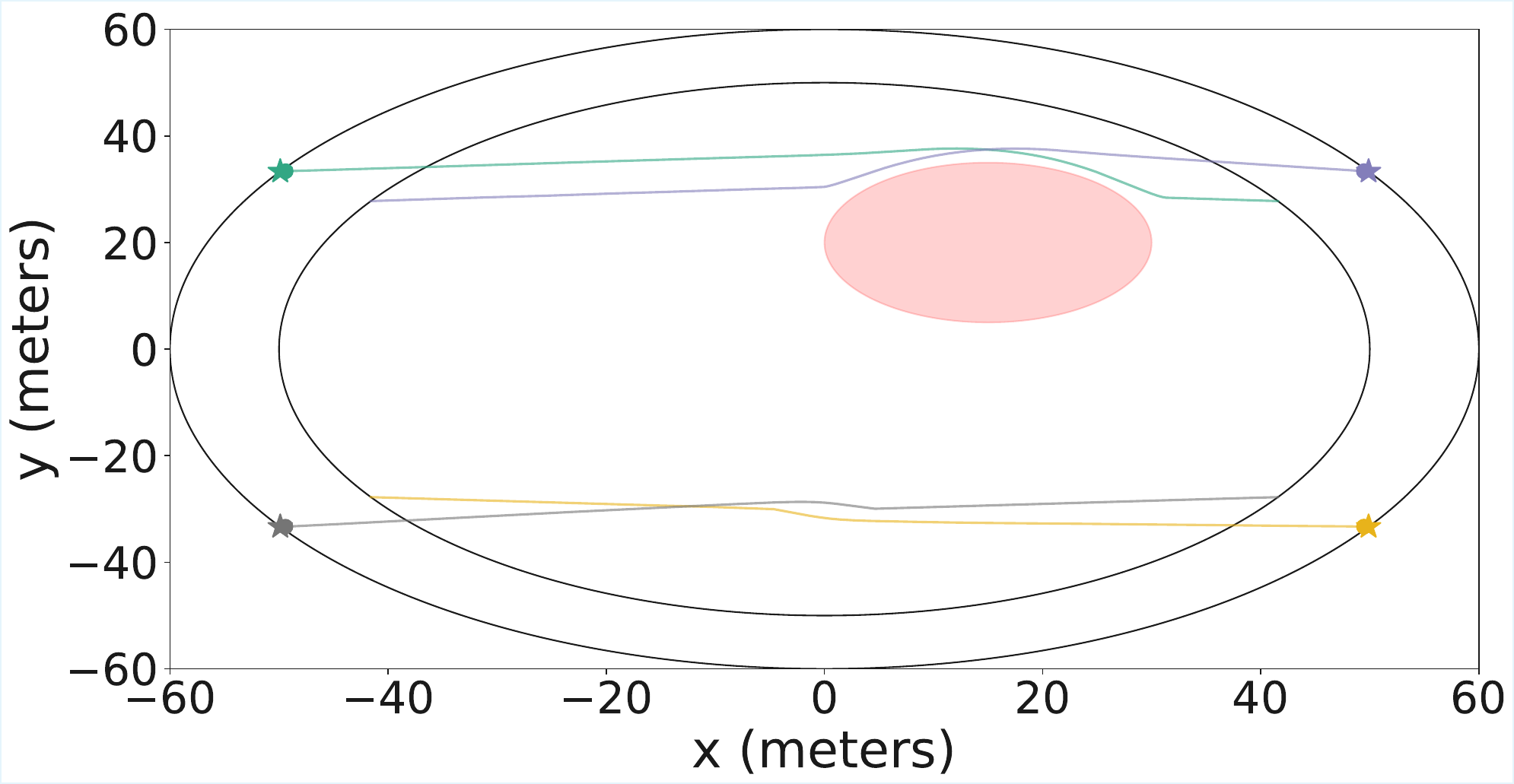}
    \caption{Static obstacle with $N=4$.}
    \label{fig:figure5}
  \end{subfigure}
  \hfill
  \begin{subfigure}{0.3\textwidth}
    \includegraphics[width=\linewidth]{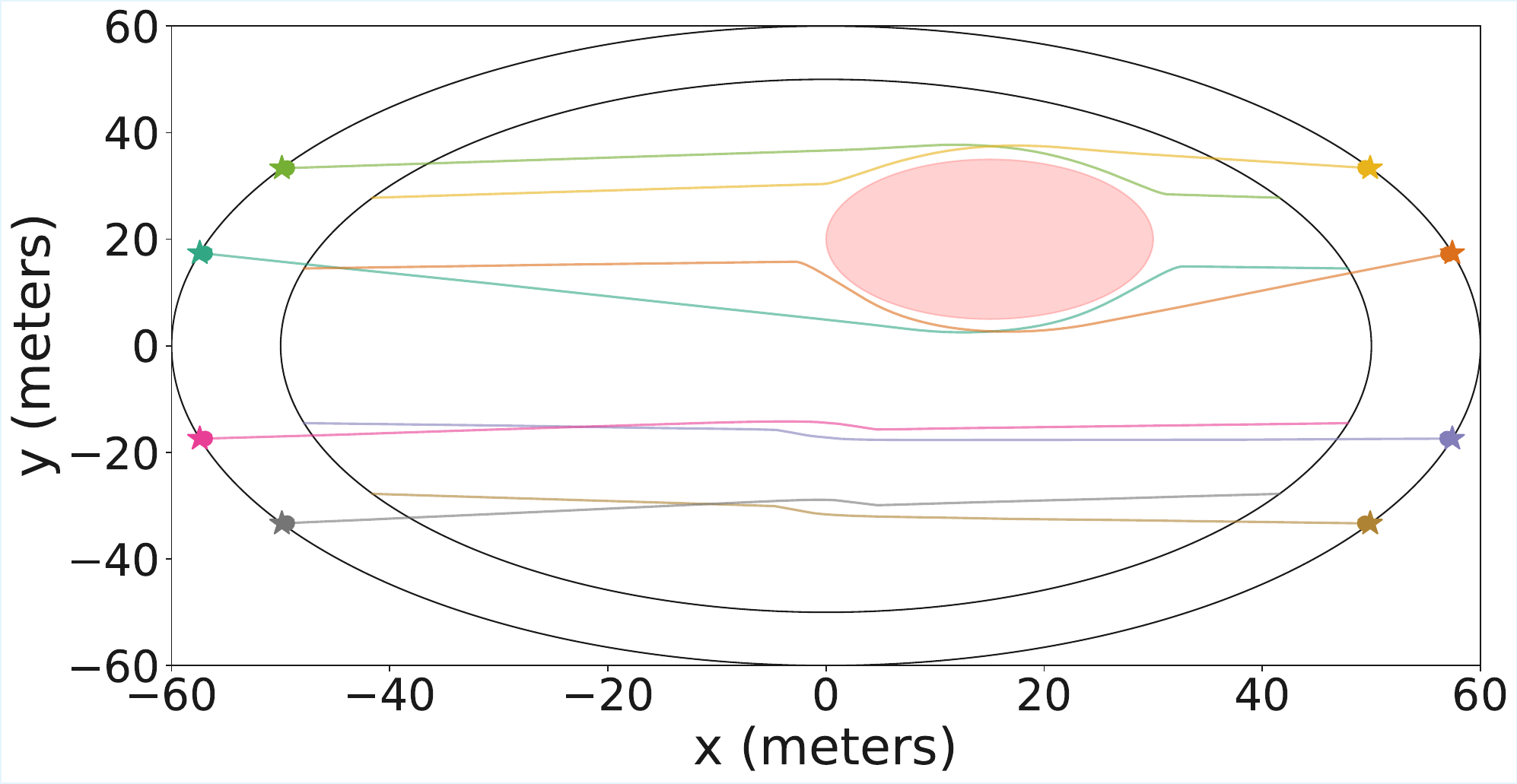}
    \caption{Static obstacle with $N=8$.}
    \label{fig:figure6}
  \end{subfigure}
  \caption{Visually demonstrating the Roundabout effect of \du in both complex and simple scenarios.}
\label{fig:resultsfig}
\end{figure*}


\subsection{The Challenge of Spatial Constraints} 

\label{Experimentation}
In Table~\ref{tab:experiment_iesults_equal}, we demonstrate the challenging nature of cluttered and constrained environments. Specifically, by removing all spatial constraints (Table~\ref{tab:experiment_iesults_equal}, \textit{row 2}), the robots are the fastest in reaching their goal, almost $3\times$ faster than in the presence of an obstacle. Additionally the robots respect a healthy distance from each other, resulting in a high MMD. On the other hand, the small gap constraint slowed the robots, but not by much. However, the robots cut too close to each other when navigating through a small gap, which is why the MMD reduces significantly.

\begin{table}[t] 
\centering 
\caption{Comparing \du with standard collision avoidance baselines. \textbf{Bold} is best.} 
\label{tab: comparison} 
\resizebox{\columnwidth}{!}{\begin{tabular}{@{}lcccccccc@{}} 
\toprule 
\multirow{2}{*}{Methods} & \multicolumn{2}{c}{TTG} & \multicolumn{2}{c}{MMD} & \multicolumn{2}{c}{FR}\\ 
\cmidrule(lr){2-3} \cmidrule(lr){4-5} \cmidrule(lr){6-7}
& $n=4$ & $n=8$ & $n=4$ & $n=8$ & $n=4$ & $n=8$ \\ 
\midrule 
Classical APF w/o LLM & 12.50 & 12.50 & 0.27 & 0.28 & 0.60 & 1.20\\ 
Classical APF & 9.80 & 12.43 & 0.19 & 0.24 & 0.80 & 1.20\\ 
MPC w/o LLM & 4.76 & 7.78 & 0.29 & 0.23 & 2.10 & 3.90\\
MPC & 7.92 & 10.51 & 0.20 & 0.21 & 1.50 & 4.10 \\  
\du w/o LLM & {2.95} & \textbf{3.00} & {0.17} & 0.22 & {2.60} & \textbf{5.10}\\
\du & \textbf{2.88} & {3.13} & 0.18 & {0.25} & \textbf{2.70} & {4.90} \\ 
\bottomrule 
\end{tabular} 
}
\end{table}

\subsection{The Role and Importance of the LLM}

We measured whether the final order in which agents reached their goals matched the final priority order after the priorities were updated during navigation. Across $100$ simulations, the LLM-enabled agents managed to reach their goals in the correct order $100\%$ of the time. This was despite the priorities changing multiple times, often drastically--an agent with the lowest priority in the beginning would be later assigned the highest priority, and it would be able to catch up and arrive first to the goal. On the other hand, as expected, without the LLM, the final ordering of the agents did not match the final priority order even once. Furthermore, when testing the role of LLM without any spatial constraints, the benefits are even more pronounced. Not only does the LLM respect the priority ordering, but it also results in much faster navigation ($>4\times$ speedup, Table~\ref{tab:experiment_iesults_equal}, \textit{rows 1 and 2}).

\subsection{The Significance of the Roundabout Effect}

The roundabout effect balances safety and efficiency. In Table~\ref{tab:experiment_iesults_equal}, we see that \du is the fastest, often by orders of more than $3\times$ (and thereby maximizing efficiency), without being both overly conservative as well as overly aggressive. This can be observed from the MMD values of \du which are neither to low nor too high - indicating that \du can balance the right amount of space to give other robots while at the same time making quick progress towards the goal. Furthermore, the roundabout effect results in the highest flow rate for \du compared to other baselines. We demonstrate the roundabout effect in Figure~\ref{fig:resultsfig}.

\subsection{Comparing $\du$ with classical APF, MPC, and RIPNA}

Figure~\ref{fig:three_images} and Table~\ref{tab: comparison} compare \du with classical APF visually and quantitatively, respectively. We observe that \du achieves a significantly lower TTG. For instance, with LLM-based dialogue and $n=8$, $\du$ is approximately $4\times$ faster than classical APF and $3.4\times$ faster than MPC. This performance advantage is consistent even in the absence of an LLM, where \du is over $4\times$ faster than classical APF and {$2.6\times$ faster} than MPC, while also achieving a {$4.25\times$ higher FR than classical APF. 
Finally, we also compared $\du$ with RIPNA. We observed that \du improved the TTG by {54.68\%} and MMD by {58.63\%}.

Classical APF takes extremely long to reach their goal (high TTG) correspondingly results in a poor FR (Table~\ref{tab: comparison}, \textit{rows 1 and 2}). The large TTG can also be correlated with the high MMD which implies that the agents are very conservative and try to steer away from other robots as much as possible, which costs time. This can also be visually confirmed in Figure~\ref{fig: apf} where the agents enter a traffic jam at the gap. MPC performs similarly poorly and repels agents and pushes them back (Figure~\ref{fig: mpc}). MPC also results in lower TTG and highest MMD on account of higher conservativeness (Table~\ref{tab: comparison}, \textit{rows 3 and 4}). In Figure~\ref{fig:three_images}, we additionally note that MPC is reactive, cutting too close to the obstacle as opposed to the obstacle, which is suboptimal. In contrast, the roundabout effect created by $\du$ results in a balance between safety and agility resulting in lower TTG and high FR.




\section{Conclusion}
\label{conclusion}
We presented \du, a new approach to priority-based collision avoidance within heterogeneous robots through LLMs in dense, constrained, and crowded airspace that contains obstacles and small-gap passages. The robots communicate with each other by querying the LLMs to generate messages and engage in the dialogue conversation in order to know each other priorities and define new priorities based upon the operations. We demonstrate our approach’s effectiveness in simulated constrained environments and compared the results with other baseline and existing algorithms.
\du can be further extended to include dynamics of the vehicle,
extend in 3D and also include hard constraints like control barrier function to ensure no vehicle enter the safe of other vehicles.

\bibliographystyle{IEEEtran}
\bibliography{root, collision_avoidance}

@inproceedings{mora2,
  author    = {Alonso Mora, J. and Breitenmoser, A. and Beardsley, P. and Siegwart, R.},
  title     = {Reciprocal collision avoidance for multiple car like robots},
  booktitle = {IEEE International Conference on Robotics and Automation},
  year      = {2012},
  pages     = {360 to 366}
}

@incollection{mora1,
  author    = {Alonso Mora, J. and Breitenmoser, A. and Rufli, M. and Beardsley, P. and Siegwart, R.},
  title     = {Optimal reciprocal collision avoidance for multiple non holonomic robots},
  booktitle = {Distributed Autonomous Robotic Systems},
  publisher = {Springer},
  year      = {2013},
  pages     = {203 to 216}
}

@inproceedings{lqr,
  author    = {Bareiss, D. and van den Berg, J.},
  title     = {Reciprocal collision avoidance for robots with linear dynamics using {LQR} obstacles},
  booktitle = {IEEE International Conference on Robotics and Automation},
  year      = {2013},
  pages     = {3847 to 3853}
}

@article{FS98,
  author    = {Fiorini, P. and Shiller, Z.},
  title     = {Motion planning in dynamic environments using velocity obstacles},
  journal   = {International Journal of Robotics Research},
  year      = {1998},
  volume    = {17},
  pages     = {760 to 772}
}

@article{ics,
  author    = {Fraichard, T. and Asama, H.},
  title     = {Inevitable collision states A step towards safer robots},
  journal   = {Advanced Robotics},
  year      = {2004},
  volume    = {18},
  number    = {10},
  pages     = {1001 to 1024}
}

@inproceedings{pvo,
  author    = {Fulgenzi, C. and Spalanzani, A. and Laugier, C.},
  title     = {Dynamic obstacle avoidance in uncertain environment combining PVOs and occupancy grid},
  booktitle = {IEEE International Conference on Robotics and Automation},
  year      = {2007},
  pages     = {1610 to 1616}
}

@inproceedings{rrvo,
  author    = {Giese, A. and Latypov, D. and Amato, N. M.},
  title     = {Reciprocally rotating velocity obstacles},
  booktitle = {IEEE International Conference on Robotics and Automation},
  year      = {2014}
}

@inproceedings{clearPath,
  author    = {Guy, S. J. and Chhugani, J. and Kim, C. and Satish, N. and Lin, M. and Manocha, D. and Dubey, P.},
  title     = {Clearpath: highly parallel collision avoidance for multi agent simulation},
  booktitle = {ACM SIGGRAPH Eurographics Symposium on Computer Animation},
  year      = {2009},
  pages     = {177 to 187}
}

@inproceedings{fvo,
  author    = {Karamouzas, I. and Guy, S. J.},
  title     = {Prioritized group navigation with formation velocity obstacles},
  booktitle = {IEEE International Conference on Robotics and Automation},
  year      = {2015},
  pages     = {5983 to 5989}
}

@inproceedings{coherence,
  author    = {Kimmel, A. and Dobson, A. and Bekris, K.},
  title     = {Maintaining team coherence under the velocity obstacle framework},
  booktitle = {International Conference on Autonomous Agents and Multiagent Systems},
  year      = {2012},
  pages     = {247 to 256}
}

@article{jia1,
  author    = {Long, P. and Liu, W. and Pan, J.},
  title     = {Deep learned collision avoidance policy for distributed multiagent navigation},
  journal   = {Robotics and Automation Letters},
  year      = {2017},
  volume    = {2},
  number    = {2},
  pages     = {656 to 663}
}

@inproceedings{ics2,
  author    = {Petti, S. and Fraichard, T.},
  title     = {Safe motion planning in dynamic environments},
  booktitle = {IEEE/RSJ International Conference on Intelligent Robots and Systems},
  year      = {2005},
  pages     = {2210 to 2215}
}

@article{cco,
  author    = {Rufli, M. and Alonso Mora, J. and Siegwart, R.},
  title     = {Reciprocal collision avoidance with motion continuity constraints},
  journal   = {IEEE Transactions on Robotics},
  year      = {2013},
  volume    = {29},
  number    = {4},
  pages     = {899 to 912}
}

@inproceedings{orcadd,
  author    = {Snape, J. and van den Berg, J. and Guy, S. J. and Manocha, D.},
  title     = {Smooth and collision free navigation for multiple robots under differential drive constraints},
  booktitle = {IEEE/RSJ International Conference on Intelligent Robots and Systems},
  year      = {2010},
  pages     = {4584 to 4589}
}

@article{hrvo,
  author    = {Snape, J. and van den Berg, J. and Guy, S. J. and Manocha, D.},
  title     = {The hybrid reciprocal velocity obstacle},
  journal   = {IEEE Transactions on Robotics},
  year      = {2011},
  volume    = {27},
  number    = {4},
  pages     = {696 to 706}
}

@incollection{orca,
  author    = {van den Berg, J. and Guy, S. J. and Lin, M. and Manocha, D.},
  title     = {Reciprocal n body collision avoidance},
  booktitle = {Robotics Research The International Symposium ISRR},
  series    = {Springer Tracts in Advanced Robotics},
  volume    = {70},
  publisher = {Springer},
  year      = {2011},
  pages     = {3 to 19}
}

@inproceedings{rvo,
  author    = {van den Berg, J. and Lin, M. and Manocha, D.},
  title     = {Reciprocal velocity obstacles for real time multi agent navigation},
  booktitle = {IEEE International Conference on Robotics and Automation},
  year      = {2008},
  pages     = {1928 to 1935}
}

@inproceedings{avo,
  author    = {van den Berg, J. and Snape, J. and Guy, S. J. and Manocha, D.},
  title     = {Reciprocal collision avoidance with acceleration velocity obstacles},
  booktitle = {IEEE International Conference on Robotics and Automation},
  year      = {2011},
  pages     = {3475 to 3482}
}

@inproceedings{gvo,
  author    = {Wilkie, D. and van den Berg, J. and Manocha, D.},
  title     = {Generalized velocity obstacles},
  booktitle = {IEEE/RSJ International Conference on Intelligent Robots and Systems},
  year      = {2009},
  pages     = {5573 to 5578}
}

@article{fan2020distributed,
  title={Distributed multi-robot collision avoidance via deep reinforcement learning for navigation in complex scenarios},
  author={Fan, Tingxiang and Long, Pinxin and Liu, Wenxi and Pan, Jia},
  journal={The International Journal of Robotics Research},
  pages={856--892},
  volume={39},
  number={7},
  year={2020},
  publisher={SAGE Publications Sage UK: London, England}
}

@article{schulman2017proximal,
  title={Proximal policy optimization algorithms},
  author={Schulman, John and Wolski, Filip and Dhariwal, Prafulla and Radford, Alec and Klimov, Oleg},
  journal={arXiv preprint arXiv:1707.06347},
  year={2017}
}

@inproceedings{sathyamoorthy2020densecavoid,
  title={Densecavoid: Real-time navigation in dense crowds using anticipatory behaviors},
  author={Sathyamoorthy, Adarsh Jagan and Liang, Jing and Patel, Utsav and Guan, Tianrui and Chandra, Rohan and Manocha, Dinesh},
  booktitle={2020 IEEE International Conference on Robotics and Automation (ICRA)},
  pages={11345--11352},
  year={2020},
  organization={IEEE}
}

@inproceedings{chen2019crowd,
  title={Crowd-robot interaction: Crowd-aware robot navigation with attention-based deep reinforcement learning},
  author={Chen, Changan and Liu, Yuejiang and Kreiss, Sven and Alahi, Alexandre},
  booktitle={2019 International Conference on Robotics and Automation (ICRA)},
  pages={6015--6022},
  year={2019},
  organization={IEEE}
}

@inproceedings{chen2020relational,
  title={Relational graph learning for crowd navigation},
  author={Chen, Changan and Hu, Sha and Nikdel, Payam and Mori, Greg and Savva, Manolis},
  booktitle={2020 IEEE/RSJ International Conference on Intelligent Robots and Systems (IROS)},
  pages={10007--10013},
  year={2020},
  organization={IEEE}
}

@inproceedings{liu2020robot,
  title={Robot Navigation in Crowded Environments Using Deep Reinforcement Learning},
  author={Liu, Lucia and Dugas, Daniel and Cesari, Gianluca and Siegwart, Roland and Dub{\'e}, Renaud},
  booktitle={IEEE/RSJ International Conference on Intelligent Robots and Systems (IROS)},
  year={2020}
}

@inproceedings{guldenring2020learning,
  title={Learning Local Planners for Human-aware Navigation in Indoor Environments},
  author={Guldenring, Ronja and G{\"o}rner, Michael and Hendrich, Norman and Jacobsen, Niels Jul and Zhang, Jianwei},
  booktitle={2020 IEEE/RSJ International Conference on Intelligent Robots and Systems (IROS)},
  pages={6053--6060},
  organization={IEEE},
  year={2020}
}

@article{chen2020robot,
  title={Robot navigation in crowds by graph convolutional networks with attention learned from human gaze},
  author={Chen, Yuying and Liu, Congcong and Shi, Bertram E and Liu, Ming},
  journal={IEEE Robotics and Automation Letters},
  volume={5},
  number={2},
  pages={2754--2761},
  year={2020},
  publisher={IEEE}
}

@inproceedings{perez2020robot,
  title={Robot navigation in constrained pedestrian environments using reinforcement learning},
  author={P{\'e}rez-D’Arpino, Claudia and Liu, Can and Goebel, Patrick and Mart{\'\i}n-Mart{\'\i}n, Roberto and Savarese, Silvio},
  booktitle={2021 IEEE International Conference on Robotics and Automation (ICRA)},
  pages={1140--1146},
  year={2021},
  organization={IEEE}
}

@inproceedings{pateldwa,
  title={DWA-RL: Dynamically feasible deep reinforcement learning policy for robot navigation among mobile obstacles},
  author={Patel, Utsav and Kumar, Nithish K Sanjeev and Sathyamoorthy, Adarsh Jagan and Manocha, Dinesh},
  booktitle={2021 IEEE International Conference on Robotics and Automation (ICRA)},
  pages={6057--6063},
  year={2021},
  organization={IEEE}
}

@inproceedings{liu2020decentralized,
  title={Decentralized structural-RNN for robot crowd navigation with deep reinforcement learning},
  author={Liu, Shuijing and Chang, Peixin and Liang, Weihang and Chakraborty, Neeloy and Driggs-Campbell, Katherine},
  booktitle={2021 IEEE International Conference on Robotics and Automation (ICRA)},
  pages={3517--3524},
  year={2021},
  organization={IEEE}
}

@inproceedings{dugas2020navrep,
  title={Navrep: Unsupervised representations for reinforcement learning of robot navigation in dynamic human environments},
  author={Dugas, Daniel and Nieto, Juan and Siegwart, Roland and Chung, Jen Jen},
  booktitle={2021 IEEE International Conference on Robotics and Automation (ICRA)},
  pages={7829--7835},
  year={2021},
  organization={IEEE}
}

@inproceedings{long2018towards,
  title={Towards optimally decentralized multi-robot collision avoidance via deep reinforcement learning},
  author={Long, Pinxin and Fan, Tingxiang and Liao, Xinyi and Liu, Wenxi and Zhang, Hao and Pan, Jia},
  booktitle={2018 IEEE International Conference on Robotics and Automation (ICRA)},
  pages={6252--6259},
  year={2018},
  organization={IEEE}
}

@article{huang2021towards,
  title={Towards Multi-Modal Perception-Based Navigation: A Deep Reinforcement Learning Method},
  author={Huang, Xueqin and Deng, Han and Zhang, Wei and Song, Ran and Li, Yibin},
  journal={IEEE Robotics and Automation Letters},
  volume={6},
  number={3},
  pages={4986--4993},
  year={2021},
  publisher={IEEE}
}

@inproceedings{everett2018motion,
  title={Motion planning among dynamic, decision-making agents with deep reinforcement learning},
  author={Everett, Michael and Chen, Yu Fan and How, Jonathan P},
  booktitle={2018 IEEE/RSJ International Conference on Intelligent Robots and Systems (IROS)},
  pages={3052--3059},
  year={2018},
  organization={IEEE}
}

@inproceedings{chen2017socially,
  title={Socially aware motion planning with deep reinforcement learning},
  author={Chen, Yu Fan and Everett, Michael and Liu, Miao and How, Jonathan P},
  booktitle={2017 IEEE/RSJ International Conference on Intelligent Robots and Systems (IROS)},
  pages={1343--1350},
  year={2017},
  organization={IEEE}
}

@article{everett2021collision,
  title={Collision avoidance in pedestrian-rich environments with deep reinforcement learning},
  author={Everett, Michael and Chen, Yu Fan and How, Jonathan P},
  journal={IEEE Access},
  volume={9},
  pages={10357--10377},
  year={2021},
  publisher={IEEE}
}

@inproceedings{xu2021human,
  title={Human-Inspired Multi-Agent Navigation using Knowledge Distillation},
  author={Xu, Pei and Karamouzas, Ioannis},
  booktitle={2021 IEEE/RSJ International Conference on Intelligent Robots and Systems (IROS)},
  pages={8105--8112},
  year={2021},
  organization={IEEE}
}

@article{chandra2025deadlock,
  title={Deadlock-free, safe, and decentralized multi-robot navigation in social mini-games via discrete-time control barrier functions},
  author={Chandra, Rohan and Zinage, Vrushabh and Bakolas, Efstathios and Stone, Peter and Biswas, Joydeep},
  journal={Autonomous Robots},
  volume={49},
  number={2},
  pages={12},
  year={2025},
  publisher={Springer}
}

@article{SUDHAKAR20201,
title = {Unmanned Aerial Vehicle (UAV) based Forest Fire Detection and monitoring for reducing false alarms in forest-fires},
journal = {Computer Communications},
volume = {149},
pages = {1-16},
year = {2020},
issn = {0140-3664},
doi = {https://doi.org/10.1016/j.comcom.2019.10.007},
url = {https://www.sciencedirect.com/science/article/pii/S0140366419308655},
author = {S. Sudhakar and V. Vijayakumar and C. {Sathiya Kumar} and V. Priya and Logesh Ravi and V. Subramaniyaswamy}
}

@INPROCEEDINGS{khatib,
  author={Khatib, O.},
  booktitle={Proceedings. 1985 IEEE International Conference on Robotics and Automation}, 
  title={Real-time obstacle avoidance for manipulators and mobile robots}, 
  year={1985},
  volume={2},
  number={},
  pages={500-505},
  doi={10.1109/ROBOT.1985.1087247}}

@article{George2009ARI,
  title={A Reactive Inverse PN algorithm for collision avoidance among multiple Unmanned Aerial Vehicles},
  author={J. M. George and Debasish Ghose},
  journal={2009 American Control Conference},
  year={2009},
  pages={3890-3895},
  url={https://api.semanticscholar.org/CorpusID:41548167}
}

@misc{openai2024gpt4ocard,
      title={GPT-4o System Card}, 
      author={OpenAI and : and Aaron Hurst and Adam Lerer and Adam P. Goucher and et. al.},
      year={2024},
      eprint={2410.21276},
      archivePrefix={arXiv},
      primaryClass={cs.CL},
      url={https://arxiv.org/abs/2410.21276}, 
}

@ARTICLE{8002674,
  author={Witwicki, Stefan and Castillo, Jose Carlos and Messias, Joao and Capitan, Jesus and Melo, Francisco S. and Lima, Pedro U. and Veloso, Manuela},
  journal={IEEE Robotics \& Automation Magazine}, 
  title={Autonomous Surveillance Robots: A Decision-Making Framework for Networked Muiltiagent Systems}, 
  year={2017},
  volume={24},
  number={3},
  pages={52-64},
  doi={10.1109/MRA.2017.2662222}}

@ARTICLE{9042827,
  author={Fischer, Juliane and Lieberoth-Leden, Christian and Fottner, Johannes and Vogel-Heuser, Birgit},
  journal={IEEE Transactions on Automation Science and Engineering}, 
  title={Design, Application, and Evaluation of a Multiagent System in the Logistics Domain}, 
  year={2020},
  volume={17},
  number={3},
  pages={1283-1296},
  doi={10.1109/TASE.2020.2979137}}

@ARTICLE{10113719,
  author={Li, Baiyu and Ma, Hang},
  journal={IEEE Robotics and Automation Letters}, 
  title={Double-Deck Multi-Agent Pickup and Delivery: Multi-Robot Rearrangement in Large-Scale Warehouses}, 
  year={2023},
  volume={8},
  number={6},
  pages={3701-3708},
  doi={10.1109/LRA.2023.3272272}}

@article{lanza2020agents,
  title={Agents and robots for collaborating and supporting physicians in healthcare scenarios},
  author={Lanza, Francesco and Seidita, Valeria and Chella, Antonio},
  journal={Journal of biomedical informatics},
  volume={108},
  pages={103483},
  year={2020},
  publisher={Elsevier}
}

@article{alotaibi2016multi,
  title={Multi-robot path-planning problem for a heavy traffic control application: A survey},
  author={Alotaibi, Ebtehal Turki Saho and Al-Rawi, Hisham},
  journal={International Journal of Advanced Computer Science and Applications},
  volume={7},
  number={6},
  year={2016},
  publisher={Science and Information (SAI) Organization Limited}
}

@inproceedings{hansen2004dynamic,
  title={Dynamic programming for partially observable stochastic games},
  author={Hansen, Eric A and Bernstein, Daniel S and Zilberstein, Shlomo},
  booktitle={AAAI},
  volume={4},
  pages={709--715},
  year={2004}
}

@misc{chandra2025deadlockfreesafedecentralizedmultirobot,
      title={Deadlock-free, Safe, and Decentralized Multi-Robot Navigation in Social Mini-Games via Discrete-Time Control Barrier Functions}, 
      author={Rohan Chandra and Vrushabh Zinage and Efstathios Bakolas and Peter Stone and Joydeep Biswas},
      year={2025},
      eprint={2308.10966},
      archivePrefix={arXiv},
      primaryClass={cs.RO},
      url={https://arxiv.org/abs/2308.10966}, 
}

@inproceedings{chen2024scalable,
  title={Scalable multi-robot collaboration with large language models: Centralized or decentralized systems?},
  author={Chen, Yongchao and Arkin, Jacob and Zhang, Yang and Roy, Nicholas and Fan, Chuchu},
  booktitle={2024 IEEE International Conference on Robotics and Automation (ICRA)},
  pages={4311--4317},
  year={2024},
  organization={IEEE}
}

@inproceedings{mandi2024roco,
  title={Roco: Dialectic multi-robot collaboration with large language models},
  author={Mandi, Zhao and Jain, Shreeya and Song, Shuran},
  booktitle={2024 IEEE International Conference on Robotics and Automation (ICRA)},
  pages={286--299},
  year={2024},
  organization={IEEE}
}

@article{gielis2022critical,
  title={A critical review of communications in multi-robot systems},
  author={Gielis, Jennifer and Shankar, Ajay and Prorok, Amanda},
  journal={Current robotics reports},
  volume={3},
  number={4},
  pages={213--225},
  year={2022},
  publisher={Springer}
}

@article{chen2024emos,
  title={Emos: Embodiment-aware heterogeneous multi-robot operating system with llm agents},
  author={Chen, Junting and Yu, Checheng and Zhou, Xunzhe and Xu, Tianqi and Mu, Yao and Hu, Mengkang and Shao, Wenqi and Wang, Yikai and Li, Guohao and Shao, Lin},
  journal={arXiv preprint arXiv:2410.22662},
  year={2024}
}

@inproceedings{liu2025coherent,
  title={Coherent: Collaboration of heterogeneous multi-robot system with large language models},
  author={Liu, Kehui and Tang, Zixin and Wang, Dong and Wang, Zhigang and Li, Xuelong and Zhao, Bin},
  booktitle={2025 IEEE International Conference on Robotics and Automation (ICRA)},
  pages={10208--10214},
  year={2025},
  organization={IEEE}
}

@article{wang2024dart,
  title={Dart-llm: Dependency-aware multi-robot task decomposition and execution using large language models},
  author={Wang, Yongdong and Xiao, Runze and Kasahara, Jun Younes Louhi and Yajima, Ryosuke and Nagatani, Keiji and Yamashita, Atsushi and Asama, Hajime},
  journal={arXiv preprint arXiv:2411.09022},
  year={2024}
}

@article{yu2023co,
  title={Co-navgpt: Multi-robot cooperative visual semantic navigation using large language models},
  author={Yu, Bangguo and Kasaei, Hamidreza and Cao, Ming},
  journal={arXiv preprint arXiv:2310.07937},
  year={2023}
}

@article{chandra2022gameplan,
  title={Gameplan: Game-theoretic multi-agent planning with human drivers at intersections, roundabouts, and merging},
  author={Chandra, Rohan and Manocha, Dinesh},
  journal={IEEE Robotics and Automation Letters},
  volume={7},
  number={2},
  pages={2676--2683},
  year={2022},
  publisher={IEEE}
}

@inproceedings{suriyarachchi2022gameopt,
  title={GAMEOPT: Optimal Real-time Multi-Agent Planning and Control at Dynamic Intersections},
  author={Suriyarachchi, Nilesh and Chandra, Rohan and Baras, John S and Manocha, Dinesh},
  booktitle={2022 IEEE 25th International Conference on Intelligent Transportation Systems (ITSC)},
  pages={2599--2606},
  year={2022},
  organization={IEEE doi - 10.1109/ITSC55140.2022.9921968}
}

@inproceedings{chandra2022game,
  title={Game-Theoretic Planning for Autonomous Driving among Risk-Aware Human Drivers},
  author={Chandra, Rohan and Wang, Mingyu and Schwager, Mac and Manocha, Dinesh},
  booktitle={2022 International Conference on Robotics and Automation (ICRA)},
  pages={2876--2883},
  year={2022}
}

@article{chandra2023socialmapf,
  title={Socialmapf: Optimal and efficient multi-agent path finding with strategic agents for social navigation},
  author={Chandra, Rohan and Maligi, Rahul and Anantula, Arya and Biswas, Joydeep},
  journal={IEEE Robotics and Automation Letters},
  volume={8},
  number={6},
  pages={3214--3221},
  year={2023},
  publisher={IEEE}
}

@article{francis2025principles,
  title={Principles and guidelines for evaluating social robot navigation algorithms},
  author={Francis, Anthony and P{\'e}rez-d’Arpino, Claudia and Li, Chengshu and Xia, Fei and Alahi, Alexandre and Alami, Rachid and Bera, Aniket and Biswas, Abhijat and Biswas, Joydeep and Chandra, Rohan and others},
  journal={ACM Transactions on Human-Robot Interaction},
  volume={14},
  number={2},
  pages={1--65},
  year={2025},
  publisher={ACM New York, NY}
}

@inproceedings{wu2023intent,
  title={Intent-aware planning in heterogeneous traffic via distributed multi-agent reinforcement learning},
  author={Wu, Xiyang and Chandra, Rohan and Guan, Tianrui and Bedi, Amrit and Manocha, Dinesh},
  booktitle={Conference on Robot Learning},
  pages={446--477},
  year={2023},
  organization={PMLR}
}

@inproceedings{chandra2024socialgym,
  title={SOCIALGYM 2.0: simulator for multi-robot learning and navigation in shared human spaces},
  author={Chandra, Rohan and Sprague, Zayne and Biswas, Joydeep},
  booktitle={Proceedings of the AAAI Conference on Artificial Intelligence},
  volume={38},
  number={21},
  pages={23778--23780},
  year={2024}
}

@inproceedings{raj2024rethinking,
  title={Rethinking social robot navigation: Leveraging the best of two worlds},
  author={Raj, Amir Hossain and Hu, Zichao and Karnan, Haresh and Chandra, Rohan and Payandeh, Amirreza and Mao, Luisa and Stone, Peter and Biswas, Joydeep and Xiao, Xuesu},
  booktitle={2024 IEEE International Conference on Robotics and Automation (ICRA)},
  pages={16330--16337},
  year={2024},
  organization={IEEE}
}

@article{suriyarachchi2024gameopt+,
  title={Gameopt+: Improving fuel efficiency in unregulated heterogeneous traffic intersections via optimal multi-agent cooperative control},
  author={Suriyarachchi, Nilesh and Chandra, Rohan and Anantula, Arya and Baras, John S and Manocha, Dinesh},
  journal={arXiv preprint arXiv:2405.16430},
  year={2024}
}

@inproceedings{zinage2025decentralized,
  title={Decentralized safe and scalable multi-agent control under limited actuation},
  author={Zinage, Vrushabh and Jha, Abhishek and Chandra, Rohan and Bakolas, Efstathios},
  booktitle={2025 IEEE International Conference on Robotics and Automation (ICRA)},
  pages={12105--12111},
  year={2025},
  organization={IEEE}
}

@article{gouru2024livenet,
  title={Livenet: Robust, minimally invasive multi-robot control for safe and live navigation in constrained environments},
  author={Gouru, Srikar and Lakkoju, Siddharth and Chandra, Rohan},
  journal={arXiv preprint arXiv:2412.04659},
  year={2024}
}

@article{song2025group,
  title={Group Fairness in Multi-Task Reinforcement Learning},
  author={Song, Kefan and Jiang, Runnan and Chandra, Rohan and Zhang, Shangtong},
  journal={arXiv preprint arXiv:2503.07817},
  year={2025}
}

@article{mahadevan2025gamechat,
  title={Gamechat: Multi-llm dialogue for safe, agile, and socially optimal multi-agent navigation in constrained environments},
  author={Mahadevan, Vagul and Zhang, Shangtong and Chandra, Rohan},
  journal={arXiv preprint arXiv:2503.12333},
  year={2025}
}

@article{chen2025livepoint,
  title={LIVEPOINT: Fully Decentralized, Safe, Deadlock-Free Multi-Robot Control in Cluttered Environments with High-Dimensional Inputs},
  author={Chen, Jeffrey and Chandra, Rohan},
  journal={arXiv preprint arXiv:2503.13098},
  year={2025}
}

@article{song2025reward,
  title={Reward Is Enough: LLMs Are In-Context Reinforcement Learners},
  author={Song, Kefan and Moeini, Amir and Wang, Peng and Gong, Lei and Chandra, Rohan and Qi, Yanjun and Zhang, Shangtong},
  journal={arXiv preprint arXiv:2506.06303},
  year={2025}
}

@article{chandra2025multi,
  title={Multi-robot navigation in social mini-games: Definitions, taxonomy, and algorithms},
  author={Chandra, Rohan and Singh, Shubham and Luo, Wenhao and Sycara, Katia},
  journal={arXiv preprint arXiv:2508.13459},
  year={2025}
}

@inproceedings{chandra2019robusttp,
  title={Robusttp: End-to-end trajectory prediction for heterogeneous road-agents in dense traffic with noisy sensor inputs},
  author={Chandra, Rohan and Bhattacharya, Uttaran and Roncal, Christian and Bera, Aniket and Manocha, Dinesh},
  booktitle={Proceedings of the 3rd ACM Computer Science in Cars Symposium},
  pages={1--9},
  year={2019}
}

@inproceedings{chandra2019densepeds,
  title={Densepeds: Pedestrian tracking in dense crowds using front-rvo and sparse features},
  author={Chandra, Rohan and Bhattacharya, Uttaran and Bera, Aniket and Manocha, Dinesh},
  booktitle={2019 IEEE/RSJ International Conference on Intelligent Robots and Systems (IROS)},
  pages={468--475},
  year={2019},
  organization={IEEE}
}

@article{chandra2020forecasting,
  title={Forecasting trajectory and behavior of road-agents using spectral clustering in graph-lstms},
  author={Chandra, Rohan and Guan, Tianrui and Panuganti, Srujan and Mittal, Trisha and Bhattacharya, Uttaran and Bera, Aniket and Manocha, Dinesh},
  journal={IEEE Robotics and Automation Letters},
  volume={5},
  number={3},
  pages={4882--4890},
  year={2020},
  publisher={IEEE}
}

@inproceedings{chandra2020graphrqi,
  title={Graphrqi: Classifying driver behaviors using graph spectrums},
  author={Chandra, Rohan and Bhattacharya, Uttaran and Mittal, Trisha and Li, Xiaoyu and Bera, Aniket and Manocha, Dinesh},
  booktitle={2020 IEEE International Conference on Robotics and Automation (ICRA)},
  pages={4350--4357},
  year={2020},
  organization={IEEE}
}

@inproceedings{chandra2020cmetric,
  title={Cmetric: A driving behavior measure using centrality functions},
  author={Chandra, Rohan and Bhattacharya, Uttaran and Mittal, Trisha and Bera, Aniket and Manocha, Dinesh},
  booktitle={2020 IEEE/RSJ International Conference on Intelligent Robots and Systems (IROS)},
  pages={2035--2042},
  year={2020},
  organization={IEEE}
}

@inproceedings{chandra2020roadtrack,
  title={RoadTrack: Realtime Tracking of Road Agents in Dense and Heterogeneous Environments},
  author={Chandra, Rohan and Bhattacharya, Uttaran and Randhavane, Tanmay and Bera, Aniket and Manocha, Dinesh},
  booktitle={2020 IEEE International Conference on Robotics and Automation (ICRA)},
  pages={1270--1277},
  year={2020},
  organization={IEEE}
}

@inproceedings{chandra2019traphic,
  title={Traphic: Trajectory prediction in dense and heterogeneous traffic using weighted interactions},
  author={Chandra, Rohan and Bhattacharya, Uttaran and Bera, Aniket and Manocha, Dinesh},
  booktitle={Proceedings of the IEEE/CVF Conference on Computer Vision and Pattern Recognition},
  pages={8483--8492},
  year={2019}
}

@phdthesis{chandra2022towards,
  title={Towards autonomous driving in dense, heterogeneous, and unstructured traffic},
  author={Chandra, Rohan},
  year={2022},
  school={University of Maryland, College Park}
}

@article{chandra2021using,
  title={Using graph-theoretic machine learning to predict human driver behavior},
  author={Chandra, Rohan and Bera, Aniket and Manocha, Dinesh},
  journal={IEEE Transactions on Intelligent Transportation Systems},
  volume={23},
  number={3},
  pages={2572--2585},
  year={2021},
  publisher={IEEE}
}

@article{moeini2025survey,
  title={A survey of in-context reinforcement learning},
  author={Moeini, Amir and Wang, Jiuqi and Beck, Jacob and Blaser, Ethan and Whiteson, Shimon and Chandra, Rohan and Zhang, Shangtong},
  journal={arXiv preprint arXiv:2502.07978},
  year={2025}
}

\end{document}